\documentclass[journal]{IEEEtran}

\usepackage{times}
\usepackage{graphicx}
\usepackage{amsmath}
\usepackage{cases}
\usepackage{bm}
\usepackage{enumerate}
\usepackage{array}
\usepackage{amsmath,amssymb,amsfonts}
\usepackage{algorithm,algorithmic}
\usepackage{subcaption}

\usepackage{epstopdf}
\usepackage{pgf}

\usepackage{multirow, booktabs} 
\usepackage{verbatim}
\usepackage{color, soul}
\usepackage{tikz}
\usetikzlibrary{arrows,positioning,patterns}

\newcommand{\revi}[1]{\textcolor{black}{#1}}
\newcommand{\revise}[1]{\textcolor{black}{#1}} 
\newcommand{\rev}[1]{\textcolor{black}{#1}} 

\newcommand{\beginsupplement}{%
	\setcounter{table}{0}
	\renewcommand{\thetable}{A\Roman{table}}%
	\setcounter{figure}{0}
	\renewcommand{\thefigure}{A\arabic{figure}}%
	\setcounter{section}{0}
	\renewcommand{\thesection}{\Roman{section}}%
}

\ifCLASSINFOpdf
\else
\fi

\begin{document}%
\title{A Cost-Sensitive Deep Belief Network for Imbalanced Classification}
%
%
%

\author{Chong~Zhang, 
        Kay~Chen~Tan,~\IEEEmembership{Fellow,~IEEE},~
        Haizhou~Li,~\IEEEmembership{Fellow,~IEEE},~
        and Geok~Soon~Hong

\thanks{C. Zhang and H.~Li are with the Department of Electrical and Computer Engineering, G. S. Hong is with the Department of Mechanical Engineering, National University of Singapore, 4 Engineering Drive 3, 117583, Singapore (e-mail: zhangchong@u.nus.edu; haizhou.li@nus.edu.sg; mpehgs@nus.edu.sg)}
\thanks{K. C. Tan is with the Department of Computer Science, City University of Hong Kong, 83 Tat Chee Avenue, Kowloon, Hong Kong.(e-mail: kaytan@cityu.edu.hk)}
\thanks{This paper has been accepted by \textit{IEEE TRANSACTIONS ON NEURAL NETWORKS AND LEARNING SYSTEMS} in April 2018.}
}


\maketitle

\begin{abstract}
    Imbalanced data with a skewed class distribution are common in many real-world applications. Deep Belief Network (DBN) is a machine learning technique that is effective in classification tasks. However, conventional DBN does not work well for imbalanced data classification because it assumes equal costs for each class. To deal with this problem, cost-sensitive approaches assign different misclassification costs for different classes without disrupting the true data sample distributions. However, due to lack of prior knowledge, the misclassification costs are usually unknown and hard to choose in practice. Moreover, it has not been well studied as to how cost-sensitive learning could improve DBN performance on imbalanced data problems. This paper proposes an evolutionary cost-sensitive deep belief network (ECS-DBN) for imbalanced classification. ECS-DBN uses adaptive differential evolution to optimize the misclassification costs based on training data, that presents an effective approach to incorporating the evaluation measure (i.e. G-mean) into the objective function. We first optimize the misclassification costs, then apply them to deep belief network. Adaptive differential evolution optimization is implemented as the optimization algorithm that automatically updates its corresponding parameters without the need of prior domain knowledge. The experiments have shown that the proposed approach consistently outperforms the state-of-the-art on both benchmark datasets and real-world dataset for fault diagnosis in tool condition monitoring.
\end{abstract}



\section{Introduction}
\IEEEPARstart{C}{lass} imbalance with disproportionate number of class instances commonly affects the quality of learning algorithms. Multifarious \revise{imbalanced} data problems \revise{exist in} \rev{numerous} real-world applications, such as fault diagnosis~\cite{zhang2015deep}, recommendation systems, fraud detection~\cite{fawcett1997adaptive}, risk management~\cite{ezawa1996learning}, tool condition monitoring~\cite{sun2004multiclassification,zhang2017datadriven,xu2018gru} and medical diagnosis~\cite{valdovinos2005class}, brain computer interface (BCI)~\cite{goh2014artifact,goh2016multiway}, data visualization~\cite{wang2017histogram}, etc. As a result of the equal misclassification costs or balanced class distribution assumption, the traditional learning algorithms are prone to the majority class when dealing with complicated classification problems \rev{that have} \revise{skewed} class distribution. \rev{Such} imbalanced data \revise{often lead to degradation of performance in learning and classification systems}. Typically, imbalance learning \rev{can be} categorized into two conventional approaches\revise{, namely data level approaches and algorithm level approaches~\cite{he2013imbalanced}.} \revise{The typical data level approaches are based on resampling approaches~\cite{chawla2002smote,he2008adasyn,han2005borderline,garcia2009evolutionary,garci2012evolutionary,lim2016evolutionary}. Some well-known resampling based approaches include synthetic minority over-sampling technique (SMOTE)~\cite{chawla2002smote} and adaptive synthetic sampling approach (ADASYN)~\cite{he2008adasyn}, etc. SMOTE~\cite{chawla2002smote} is an oversampling technique that generates synthetic samples of minority class. ADASYN~\cite{he2008adasyn} uses a weighted distribution for different minority class according to their level of difficulty in learning and more synthetic data for minority class. A typical algorithm level approach is called cost-sensitive learning~\cite{zhou2006training,Datta2015near,Zong2013weighted,castro2013novel,Kremp2013optimized}. We focus the study of algorithm level approach in this paper.} 

Resampling approaches attempt to \revise{manually} rebalance dataset by oversampling minority samples and/or under-sampling majority samples. \rev{Unfortunately, such} approaches may\revise{, on one hand miss out} potentially useful data\rev{, on the other hand add} the computational burden with the redundant samples. Essentially, resampling-based \revise{approaches} would alter the original distribution of classes. \revise{In practice}, the assumption that all misclassification errors have equal costs is not true in real-world applications. \rev{There could be} large \rev{differences in terms of costs} between different misclassification errors. For instance, in fault diagnosis of tool condition, if \rev{we are detecting the healthy state versus failure state of a machine,} \rev{we know that} missing \rev{the detection of a failure state} may cause a catastrophic accident which costs much higher than \revise{the others}.  

Many conventional approaches presume equal costs for all the classification errors and this assumption usually \revise{does not hold} in practice. Some real-world problems have drastically various costs for different classes\revise{, for example, between} failure state and healthy state of \revise{a machine}. Cost-sensitive learnings are popular methods \revise{that deal with} imbalanced classification problems with unknown and unidentical costs on algorithmic level. The intuition of cost-sensitive learning is to assign misclassification costs for each class appropriately. \rev{We have seen studies} in cost-sensitive neural networks~\cite{zhou2006training}, cost-sensitive decision trees~\cite{drummond2003c4}, cost-sensitive extreme learning machine~\cite{Zong2013weighted}, etc. However, there are few studies \rev{on} cost-sensitive deep belief networks. Deep belief network (DBN)~\cite{hinton2006fast,hinton2010practical}, a generative model stacked with several Restricted Boltzmann Machines (RBMs), has drawn tremendous attention \revise{recently. It has shown promising results in classification tasks such as}, image identification, speech recognition~\cite{mohamed2012acoustic}, \rev{and} natural language processing~\cite{mohamed2011deep}. \revise{DBN is known for its} extraordinary end-to-end feature learning and classification characteristics. However, it has not been investigated as to how cost-sensitive learning \revise{could enhance} DBN to deal with imbalanced data problems. 

\revise{A key issue in cost-sensitive learning is to estimate the costs associated with data classes in different problems. The generic population-based evolutionary algorithms\revise{, that successfully deal} with multi-model optimization problems~\cite{qu2012differential,yang2009firefly,zhang2009jade,zhang2016multiobjective}\revise{, offers a solution to addressing the issue.} Differential evolution (DE) is a popular variant of EAs. DE optimizes an optimization problem by iteratively searching \revise{for a solution} given an evaluation metric. Basically, DE moves the candidate solutions around the search space by using simple mathematical formulae to combine the positions of existing solutions. In this way, if a new position gives improvement, an old position is replaced, otherwise, the new position is discarded. It solves non-separable multi-model (i.e. has many local optima) problems and avoids local optima. In comparison with other variants of EAs, DE has better exploration capability with fewer parameters. It is also easy to implement. However, the exploration and exploitation capabilities of DE are mainly controlled by two key parameters, i.e. the mutation factor and crossover probability. Traditional DEs use fixed parameters which are not suitable for different problems and hard to tune. Adaptive DE~\cite{zhang2009jade} automatically updates the parameters according to the probability matching, that can be easily implemented. Therefore, it becomes a logical choice in solving practical problems.}

\rev{The above observation\rev{s have} promoted us to study an} Evolutionary Cost-Sensitive Deep Belief Network (ECS-DBN) to deal with \rev{the} \revise{imbalanced} data problems, where \revise{we find ways to assign differential misclassification costs to the classes, that we also call class-dependent misclassification costs.} The misclassification costs are optimized by adaptive differential evolution (DE) algorithm~\cite{zhang2009jade}. \revise{We consider that such} a study could help to identify methods to heuristically optimize misclassification costs. In the \revise{rest} of the paper, we presume positive label for the minority class and negative label for the majority class. 

The contributions of this paper are summarized as follows.
\begin{enumerate}
	\item \rev{We formulate a novel learning algorithm for classification prediction of DBN that handles imbalanced data classification;}
	\item \rev{We show how ECS-DBN works by assigning appropriate misclassification costs and incorporating cost-sensitive learning with deep learning.}
	\item \rev{We show that ECS-DBN allows us to} determine the unknown misclassification costs \revise{without} prior domain knowledge.
	\item \revise{We show that ECS-DBN offers an effective solution of good performance to imbalanced classification problems. The proposed approach can automatically work for both binary and multiclass classification problems.}
\end{enumerate}

The rest of the paper \rev{is} organized as follows. Section~\ref{sec:lit} reviews current related literature. Section~\ref{sec:csdbn} and Section~\ref{sec:ecs-dbn} introduce cost-sensitive deep belief network and present the proposed ECS-DBN, respectively. Section~\ref{sec:benchmark_simulation} compares the proposed approach and other state-of-the-art methods on 58 benchmark datasets. Section~\ref{sec:real_world_experiment} reports the experiments on a real-world dataset of fault diagnosis in tool condition monitoring on gun drilling. \rev{Finally,} Section~\ref{sec:conclusion} concludes \rev{the discussion} and highlights some potential research directions.

\section{Literature Reviews}\label{sec:lit}
\subsection{Cost-sensitive Learning}
Cost-sensitive learning \rev{method~\cite{he2009learning} is} a learning paradigm that \revise{assigns differential misclassification costs to the classes involved in a classification task.}

Datta et al.~\cite{Datta2015near} investigated Near-Bayesian Support Vector Machines (NBSVM) for imbalanced classification with equal or unequal misclassification costs for multi-class scenario. Zong et al.~\cite{Zong2013weighted} proposed a weighted extreme learning machine (WELM) for imbalance learning. The approach benefits from \rev{the idea of }original extreme learning machine (ELM) which \rev{is} simple and convenient to implement. It can be applied directly into multiclass classification tasks. The WELM is capable of dealing with imbalanced class distribution. The weights are assigned for each example according to users' needs. Krempl et al.~\cite{Kremp2013optimized} proposed OPAL which is a fast, non-myopic and cost-sensitive probabilistic active learning approach. However, \rev{such approaches} cannot determine \rev{the} optimal misclassification loss without \rev{the need of prior domain} knowledge. Castro et al.~\cite{castro2013novel} proposed a cost-sensitive algorithm (CSMLP) using a single cost parameter to differentiate misclassification errors to improve the performances of MLPs on binary imbalanced class distributions. ABMODLEM~\cite{Napierała2015addressing} addressed imbalanced data with argument based rule learning. By using the expert knowledge, \rev{CSMLP and ABMODLEM} improve learning rules from imbalanced data. \rev{With argument based rule induction, such approaches} is convenient for domain experts to describe reasons for specific classes.

\revise{There are many studies related to neural networks over imbalanced learning. Zheng~\cite{zheng2010cost} proposed cost-sensitive boosting neural networks which incorporate the weight updating rule of boosting procedure to associate samples with misclassification costs. Bertoni et al.~\cite{bertoni2011cosnet} proposed a cost-sensitive neural network for semi-supervised learning in graphs. Cost-sensitive SVM (CS-SVM)~\cite{gu2017cross} was discussed with model selection via global minimum cross validation error. Tan et al.~\cite{tan2015evolutionary} proposed an evolutionary fuzzy ARTMAP (FAM) neural network using adaptive incremental learning method to overcome the stability-plasticity dilemma on stream imbalanced data. A cost-sensitive convolutional neural network~\cite{khan2017cost} was proposed for imbalanced image classification. Despite many studies on imbalanced learning, the potential benefits through DBN with imbalanced learning have not been fully explored yet.} 

\subsection{Evolutionary \rev{Algorithm (EA)}}\label{sec:ea}
\revise{We note that it is possible to decide misclassification costs} either \revise{by} trial and error~\revise{\cite{kukar1998cost,zhou2006training}} or EAs~\revise{\cite{tan2015evolutionary,zhang2016evolutionary,li2005cost}}. Inspired by the biological evolution process, \rev{EA is a} meta-heuristic optimization method, which attracts significant attention \revise{when learning from imbalanced data}. EA based studies on optimizing imbalanced classification \rev{can be broadly grouped into two categories.}

\revise{In the first category, one can implement} EA to optimize the dataset for training classifiers. Early \rev{studies are focused} on using EA to drive the sampling process of the training dataset. Such approaches~\cite{garcia2009evolutionary,drown2007using,zou2008svm} represent data by expressing the chromosome with binary representation. However, these approaches have poor scaling ability for large datasets as the chromosome expands proportionally with the size of the dataset, resulting in a cumbersome and time consuming evolutionary process. Recent methods (e.g.~\cite{garci2012evolutionary}) attempt to circumvent this problem by employing EA to sample smaller subsets of data to represent the imbalanced dataset. Another \rev{idea}~\cite{ghazikhani2012class} \rev{is to use} EA to carry out random under-sampling by determining the optimal regions in the \revise{sample space}. Recent trends incorporating EA into \revise{sample space} are limited to repetitive sampling-based solutions. ECO-ensemble~\cite{lim2016evolutionary} incorporates synthetic data generation within an ensemble framework optimized by EA simultaneously. Although this approach integrates EA into the whole framework from sample space to model space, the computational complexity also increases accordingly. These data-level approaches are dreadfully sensitive to the quality of imbalanced data (i.e. outliers, sparse data and small disjuncts). \rev{The use of synthetic data may change the true distribution of the original dataset therefore do more harm than good to the classifier.}

\revise{In the second category, one can implement} EA \rev{by optimizing} classifiers for imbalanced classification at algorithmic level. Such approaches~\cite{harvey2015automated} optimize the classifier in \rev{the} model space by using EA. Some studies~\cite{orriols2009evolutionary, ducange2010multi,luo2016sparse} implement EAs to enhance rule-based classifiers. Genetic programming has been utilized to acquire sets of optimized classifiers such as negative correlated learning~\cite{bhowan2013evolving, bhowan2014reusing, wang2015convex}. Perez et al.~\cite{perez2010analysis} integrates an evolutionary cooperative-competitive algorithm to obtain a set of simple and accurate radial basis function networks. Due to lack of sufficient domain knowledge, the costs of misclassification in cost-sensitive methods are usually hard to determine. \rev{In the literature, we note that the} present studies \rev{are focused very much on} traditional simple network models~\revise{\cite{zhou2006training,zhang2016evolutionary,tan2015evolutionary}}. To our best knowledge, there is no reported work \rev{on the study of EA in cost sensitive deep learning}. In this paper, we study how to estimate the misclassification costs automatically to improve the performance of cost-sensitive deep belief network.

\section{Technical Details of Deep Belief Network with Cost-sensitive Learning}
\label{sec:csdbn}
\subsection{Deep Belief Network}
Deep Belief Network (DBN) is a probabilistic generative model stacked with several Restricted Boltzmann Machines (RBMs). DBN is trained by greedy unsupervised layer-wise pre-training and discriminative supervised fine-tuning. The weight connections in DBN \rev{are} between contiguous layers\rev{, there is no connections between the hidden neurons within} the same layer. 

The fundamental building block of DBN is \rev{an} RBM which consists of one visible layer and one hidden layer. To construct \rev{a} DBN, \rev{the} hidden layer of previous RBM is regarded as the visible layer of its \rev{subsequent RBM in the deep structure.} \rev{To train a DBN, typically each RBM is pre-trained initially from the bottom to the top in a layer-wise manner and subsequently the whole network} is fine-tuned with supervised learning methods. Ultimately, the \rev{hypothesized} prediction is obtained in the output layer based on the posterior probability distribution obtained from \rev{the} penultimate layer. 

DBN is usually trained by progressively untying the weights in each layer from the weights in higher layers~\cite{hinton1995wake}. \rev{The pre-training is carried out by} alternating Gibbs sampling from the true posterior distribution over all the hidden layers \rev{between} a data sample on the visible variables and the transposed weight matrices to infer the factorial distributions over each hidden layer. \rev{All the variables in one layer are updated \rev{in parallel} via Markov chain until they reach their} stationary equilibrium distribution. The log posterior probability of the data is maximized by this training procedure. While the posterior distribution is created by the likelihood term coming from the data~\cite{hinton2006fast}. Factorial approximations is used in DBN to replace the intractable true posterior distribution. By implementing a complementary prior, the true posterior is exactly factorial. 

The posterior in each layer is approximated by a factorial distribution of independent variables within a layer given the values of the variables in the previous layer. Based on wake-sleep algorithm proposed in Hinton et al.~\cite{hinton2006reducing}, the weights on the undirected connections at the top level are learned by fitting the top-level RBM to the posterior distribution of the penultimate layer. The fine-tuning starts with a state of the top-level output layer and uses the top-down generative connections to stochastically activate each lower layer in turn. So a DBN can be viewed as an RBM that defines a prior over the top layer of hidden variables in a directed belief net, combined with a set of “recognition” weights to perform fast approximate inference.

The architecture of DBN makes \revise{it} possible to abstract higher level features through layer conformation~\cite{hinton2010practical}. Each layer of hidden variables learns to represent features that capture higher order correlations in the original input data. \rev{Applying DBNs to a classification problem}, feature vectors from data samples are used to set the states of the visible variables of the lower layer of the DBN. \rev{The DBN is then trained to produce} a probability distribution over the possible labels of the data based on posterior probability distribution of the data samples.

\begin{figure}[t]
\centering
\includegraphics[width=3.0in]{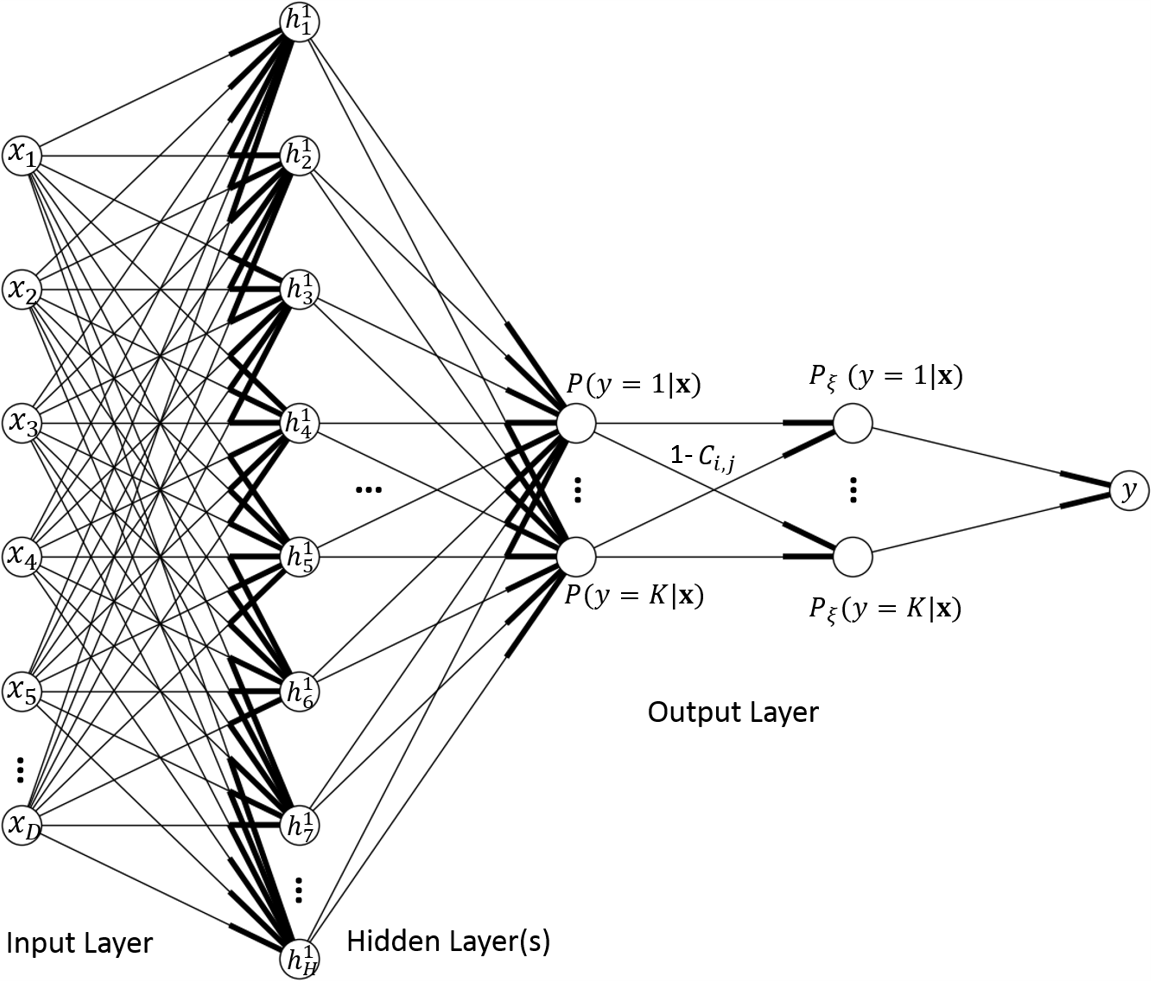}
\caption{ Schematic diagram of a cost-sensitive deep belief network (CSDBN) \rev{with misclassification costs for different classes in the output layer}.}
\label{fig:cost_sensitiveDBN}

\end{figure}

\rev{Suppose a dataset $S=\{\{\mathbf{x}_1,y_1\}, \{\mathbf{x}_2,y_2\},\cdots,\{\mathbf{x}_N,y_N\}\}$ contains a total number of $N$ data sample pairs $\{\mathbf{x}_n,y_n\}$, where $\mathbf{x}_n$ is the $n$th data sample, $y_n$ is the corresponding $n$th target label.} Assume a DBN consists of $H$ hidden layer and the parameters of each layer $i\in \{1, \cdots, H\}$ by $\theta_i=\{\mathbf{W}_i,\mathbf{b}_i\}$. Given an input data sample $\mathbf{x}$ from the dataset, the DBN with $H$ hidden layer(s) presents a complex feature mapping function. After feature transformation, softmax layer \rev{serves} as the output layer of DBN to perform classification predictions \rev{as} parameterized by $\theta_s=\{\mathbf{W}_s,\mathbf{b}_s\}$. Suppose there are $K$ neurons in the softmax layer, where the $j$-th neuron is responsible for estimating the prediction probability of class $j$ given input of \revise{$\revise{\mathbf{x}_H}$ which is the output of the previous layer} and \rev{associated} with weights $\mathbf{W}_s^{(j)}$ and bias $\mathbf{b}_s^{(j)}$：
\begin{equation}
 P(y=j|\mathbf{x})=\frac{exp(\mathbf{b}_s^{(j)}+\revise{\mathbf{x}_H}^T\mathbf{W}_s^{(j)})}{\sum_{k=1}^K exp(\mathbf{b}_s^{(k)}+\revise{\mathbf{x}_H}^T\mathbf{W}_s^{(k)})},
\end{equation}
where \revise{$\revise{\mathbf{x}_H}$ is the output of the previous layer}. Based on the probability estimation, the trained DBN classifier \rev{provides a prediction} as 
\begin{equation}
 f(\mathbf{x})=\arg\max_{1\leq j \leq K}P(y=j|\mathbf{x}). 
\end{equation}
In \rev{practice}, the parameters $\{\theta_1, \theta_2, \cdots, \theta_H, \theta_s\}$ of DBN are massively optimized by statistic gradient descent with respect to the negative log-likelihood loss over the training set $S_t$.

\subsection{Cost-sensitive Deep Belief Network}
The concept of cost-sensitive learning is to minimize the overall cost (e.g. Bayes conditional risk~\cite{elkan2001foundations}) on the training data set.

Assume the total number of classes is $K$, given a sample data $\mathbf{x}$, $C_{i,j}\in[0,1]$ denotes the cost of misclassifying $x$ as class $j$ when $x$ actually belongs to class $i$. In addition, $C_{i,j}=0$, when $i=j$, which indicates the cost for correct classification is 0. 

Given the misclassification costs \rev{$C_{i,j}$}, a data sample should be classified into the class that has the minimum expected cost. Based on decision theory~\cite{berger2013statistical}, the decision rule minimizing the expectation cost $\mathcal{R}(i|\mathbf{x})$ of classifying an input vector $\mathbf{x}$ into class $i$ can be expressed as:
\begin{equation}
 \mathcal{R}(i|\mathbf{x})=\sum_{j=1, j\neq i}^K P(j|\mathbf{x})C_{i,j}~,
\end{equation}
where $P(j|\mathbf{x})$ is the posterior probability estimation of classifying a data sample into class $j$. 
\revise{Given the prior probability $P(\mathbf{x}_n)$, the} general decision rule indicates which action to take for each data sample $\mathbf{x}_n$, thus the overall risk $\mathcal{R}$ is
\begin{equation}
 \mathcal{R}= \sum_{n=1}^N\sum_{i=1}^K\mathcal{R}(i|\mathbf{x}_n)P(\mathbf{x}_n)
\end{equation}

According to the Bayes decision theory, an ideal classifier will give a decision by computing the expectation risk of classifying an input to each class and predicts the label that reaches the minimum overall expectation risk. Misclassification costs represent penalties for classification errors. In cost-sensitive learning, all misclassification costs are essentially non-negative.

Mathematically, the probability that a sample data $\mathbf{x} \in S$ belongs to a class $j$, a value of a stochastic variable $y$, can be expressed as :
\begin{equation}
\begin{array}{rcl}
\label{eq:estimate_prob}
 P(y=j|\mathbf{x}) & = & softmax_j(\mathbf{b}+\mathbf{W}\mathbf{x})\\
& = & \frac{exp({\mathbf{b}_j+\mathbf{W}_j x})}{\sum_i exp({\mathbf{b}_i+\mathbf{W}_ix})},
\end{array}
\end{equation}
\revi{The misclassification threshold values are introduced to turn posterior probabilities into class labels such that the misclassification costs are minimized.} \rev{Implementing} the misclassification \revi{threshold value $1-C_{i,j}$} on the obtained posterior probability $P(y=j|\mathbf{x})$, \rev{one} can obtain the \revi{new probability $P_{\xi}$}:
\begin{equation}
\label{eq:cost_f}
 P_{\xi}(y=j|\mathbf{x})=\revise{P(y=j|\mathbf{x})}\cdot (1-C_{i,j}).
\end{equation}
Generally, the misclassification threshold values for minority classes are larger than those of majority classes. The \rev{hypothesized} prediction \rev{$f(\mathbf{x})$} of the sample \rev{$\mathbf{x}$} is the member of the maximum probability among classes, can be obtained by using the following equation:
\begin{equation}
\label{eq:prediction}
 F(\mathbf{x}) = arg\max_j P_{\xi}(y=j|\mathbf{x}).
\end{equation} 

The proposed cost-sensitive learning method only concerns the output layer of a DBN. \revise{In this paper, we follow the same pre-training and fine-tuning procedures as in~\cite{hinton2006fast}}. 

For imbalanced classification problems, the prior probability distribution \revise{of} different classes is essentially imbalanced or non-uniform. To reflect class imbalance, \rev{there is a need to introduce the misclassification cost at the output layer to reflect the} imbalanced class distributions. In addition, traditional training algorithms generally assume uniform class distribution~with equal misclassification costs, i.e. $\forall i,j \in [1, 2, \cdots, K]$, if $i=j, C_{i,j}=0$, if $i \neq j, C_{i,j}=1$, 　which is not true in many real-world applications. 

In \revise{many} real-world applications, the misclassification costs are essentially unknown and they vary across various classes. \rev{The current studies}~\cite{elkan2001foundations} usually attempt to determine misclassification costs by \rev{trial} and error which \revise{generally does not lead to an optimal solution}. Some studies~\cite{kukar1998cost} have \rev{devised} mechanisms to update misclassification costs based on the number of samples in different classes. However, \rev{such} methods may not \rev{be} suitable for \rev{the} cases where \rev{the} classes are important but rare, such as some rare fatal diseases. To avoid hand tuning of misclassification costs, adaptive differential evolution algorithm~\cite{zhang2009jade} \revise{is} implemented in this paper. Adaptive differential evolution algorithm is a simple effective and efficient evolutionary algorithm which could obtain optimal solution by evolving and updating a population of individuals during several generations. It \rev{attempts to adaptively self-update the} control parameters without \rev{the need of prior} knowledge.

\section{Evolutionary Cost-Sensitive Deep Belief Network (ECS-DBN)}\label{sec:ecs-dbn}
\rev{As discussed in Section~\ref{sec:ea}, evolutionary algorithm (EA) is a widely used optimization algorithm which \rev{is motivated by} the biological evolution process. \rev{The EA algorithm can be designed to optimize} the misclassification costs \rev{that are unknown in practice.} \rev{In this paper, we propose an}} Evolutionary Cost-Sensitive Deep Belief Network (ECS-DBN) \rev{by} \rev{incorporating} cost-sensitive function directly into its classification paradigm \rev{with the misclassification costs being optimized through adaptive differential evolution~\cite{zhang2009jade,qin2005self}.} \rev{The main idea of this cost-sensitive learning technique is to assign \revise{class-dependent costs}}. Fig.~\ref{fig:cost_sensitiveDBN} shows the schematic diagram of a cost-sensitive deep belief network. 

\rev{The} procedure of training the proposed ECS-DBN \rev{can be summarized} in Table~\ref{tab:algol}. Firstly, a population of misclassification costs \rev{are \revise{randomly} initialized}. \rev{We then} \rev{train} a DBN with the training dataset. After \rev{applying} misclassification costs on the outputs of the \rev{DBN}, \rev{we evaluate} the training error based on the performance of the corresponding cost-sensitive \rev{hypothesized} prediction. According to the evaluation performance on training \rev{dataset}, proper misclassification costs \rev{are selected} to generate the population of next generation. In the next generation, mutation and crossover operators \rev{are employed} to evolve a new population of misclassification costs. Adaptive \rev{differential evolution (DE) algorithm} will proceed to next generation and continuously \rev{iterate between mutation and} selection \rev{to reach the maximum number of generations}. Eventually, the best found misclassification costs \rev{are obtained} and \rev{applied} \rev{to} the output layer of DBN to form ECS-DBN as shown in Fig.~\ref{fig:cost_sensitiveDBN}. \rev{At run-time, we test} the \rev{resulting} \rev{ECS-DBN} with \rev{test} \rev{dataset} \rev{to report} the performance. \revise{The practical steps of ECS-DBN is summarized in Algorithm~\ref{alg:ECS-DBN}, and discussed next.}

\begin{table}[t]
\caption{ \rev{The training process of an} Evolutionary Cost-Sensitive Deep Belief Network (ECS-DBN).}
\label{tab:algol}
\resizebox{\columnwidth}{!}{
\begin{tabular}{l}
\hline
\hline
\textbf{Pre-Training Phase:}                                                                                                                                              \\
1. Let $S_{t}$ be the training set.                                                                                                                                    \\
\begin{tabular}[c]{@{}l@{}}2. Pre-train DBN using greedy layer-wise training algorithm with $S_{t}$.\end{tabular}                                                  \\
\textbf{Fine-Tuning Phase:}                                                                                                                                               \\
\begin{tabular}[c]{@{}l@{}}1. Randomly initialize a population of misclassification costs.\end{tabular}                                                                          \\
\begin{tabular}[c]{@{}l@{}}2. Generate a new population of misclassification costs via mutation and \\crossover based on differential operator.\end{tabular}                       \\
\begin{tabular}[c]{@{}l@{}}3. Multiply the corresponding misclassification costs on training output\\and evaluate the error on training data.\end{tabular}                                       \\
\begin{tabular}[c]{@{}l@{}}4. According to evaluation performance, select appropriate misclassification\\costs and discard inappropriate ones in order to evolve next generation.\end{tabular} \\
\begin{tabular}[c]{@{}l@{}}5. Continuously iterate between mutation and selection to reach the\\maximum number of generations.\end{tabular}                                        \\
\hline
\end{tabular}
}

\end{table}

\begin{algorithm}[t]
	\caption{Procedure of ECS-DBN}  
	\label{alg:ECS-DBN} 
	\begin{small}
\begin{algorithmic}
\REQUIRE ~~\\
$\mathbf{X}$: Imbalanced data samples\\
$Y$: Class labels\\
$N$: Population size\\
$G$: Maximum number of generations (i.e. stopping criterion)\\
$F$: Mutation factor\\
$Cr$: Crossover probability\\		
$D$: Dimension of solution space\\
$range=(\mathbf{c}_{min},\mathbf{c}_{max})$: Range of values for chromosome.\\
Set $\mu_{Cr}=0.5,\mu_F=0.5, A=\emptyset, \beta=0.5.$\\

\STATE \textbf{Step 1) Initialization:}
Generate an initial population $\{\mathbf{c}_0^1, \dots, \mathbf{c}_0^{N}\}$ via uniformly random sampling in solution space. The initial value of $i$th individual is generated as $\mathbf{c}_0^i=\mathbf{c}_{min}+rand(0,1)\cdot (\mathbf{c}_{max}-\mathbf{c}_{min})$. And evaluate each candidate solution $\mathbf{c}_0^i$ ($i = 1,\cdots, N$) in the initial population via its corresponding trained $DBN(\mathbf{x}\in S_{t},y|\mathbf{c})$ \revise{to obtain a vector representing the \revi{fitness} functions} $F(\mathbf{c}_0^i)$ which is G-mean of training dataset in this paper.
\STATE \textbf{Step 2) Evolution:}
\FOR {$g = 1, \cdots, G$}
\STATE Set the set of all successful mutation factor $F_i$ at each generation $S_F=\emptyset;$
\STATE Set the set of all successful crossover probabilities $Cr_i$ at each generation $S_{Cr}=\emptyset;$
\FOR {$i = 1, \cdots, N_p$}
\STATE Generate $Cr_i=randn(\mu_{Cr},0.1), F_i=randc_i(\mu_F,0.1)$.
\STATE \textbf{Step 2.1) Mutation:} Randomly select two indices $j$ and $k$ from population, then generate a new candidate solution $\mathbf{c}_g^{i'}$ from $\mathbf{c}_g^{i}$, $\mathbf{c}_g^{j}$ and $\mathbf{c}_g^{k}$ by $\mathbf{c}_g^{i'}$=$\mathbf{c}_g^{i} + F_i \cdot (\mathbf{c}_g^{j} - \mathbf{c}_g^{k})$, \rev{which is a DE operator.}\\
Generate $i_{rand}=randint(1,D)$
\STATE \textbf{Step 2.2) Crossover:} 
\IF{$i=i_{rand}$ or $rand(0,1)<Cr_i$}
\STATE $\mathbf{u}_{g}^i=\mathbf{c}_{g}^{i'}$
\ELSE
\STATE $\mathbf{u}_{g}^i=\mathbf{c}_{g}^i$
\ENDIF
\STATE \textbf{Step 2.3) Selection:}
\IF{$F(\mathbf{c}_g^i)\geq F(\mathbf{u}_g^i)$}
\STATE $\mathbf{c}_{g+1}^i=\mathbf{c}_{g}^i$
\ELSE
\STATE $\mathbf{c}_{g+1}^i=\mathbf{u}_{g}^i, \mathbf{c}_{g}^i\rightarrow A, Cr_i \rightarrow S_{Cr}, F_i \rightarrow S_F$
\ENDIF
\STATE Randomly remove solutions from A so that $|A|\leq N$
\STATE \textbf{Step 2.4) Parameter Adaptation:} 
\STATE $\mu_{Cr}=(1-\beta)\cdot \mu_{Cr}+\beta\cdot mean(S_{CR})$
\STATE $\mu_F=(1-\beta)\cdot \mu_F+\beta\cdot mean(S_F)$
\ENDFOR
\ENDFOR
\STATE \textbf{Step 3) ECS-DBN Creation:} Generate an ECS-DBN with \revise{the best} individual $\mathbf{c}_{best}$ \revise{obtained from the} training dataset $S_{t}$ as the misclassification cost.
\STATE \textbf{\rev{Step 4) Run-time Evaluation:}} \rev{\revise{Evaluate} ECS-DBN on test} dataset $S_{test}$.
\end{algorithmic}
\end{small}
\end{algorithm}

\subsection{Chromosome Encoding}
\begin{figure}[t]
\centering
\includegraphics[width=3.5in]{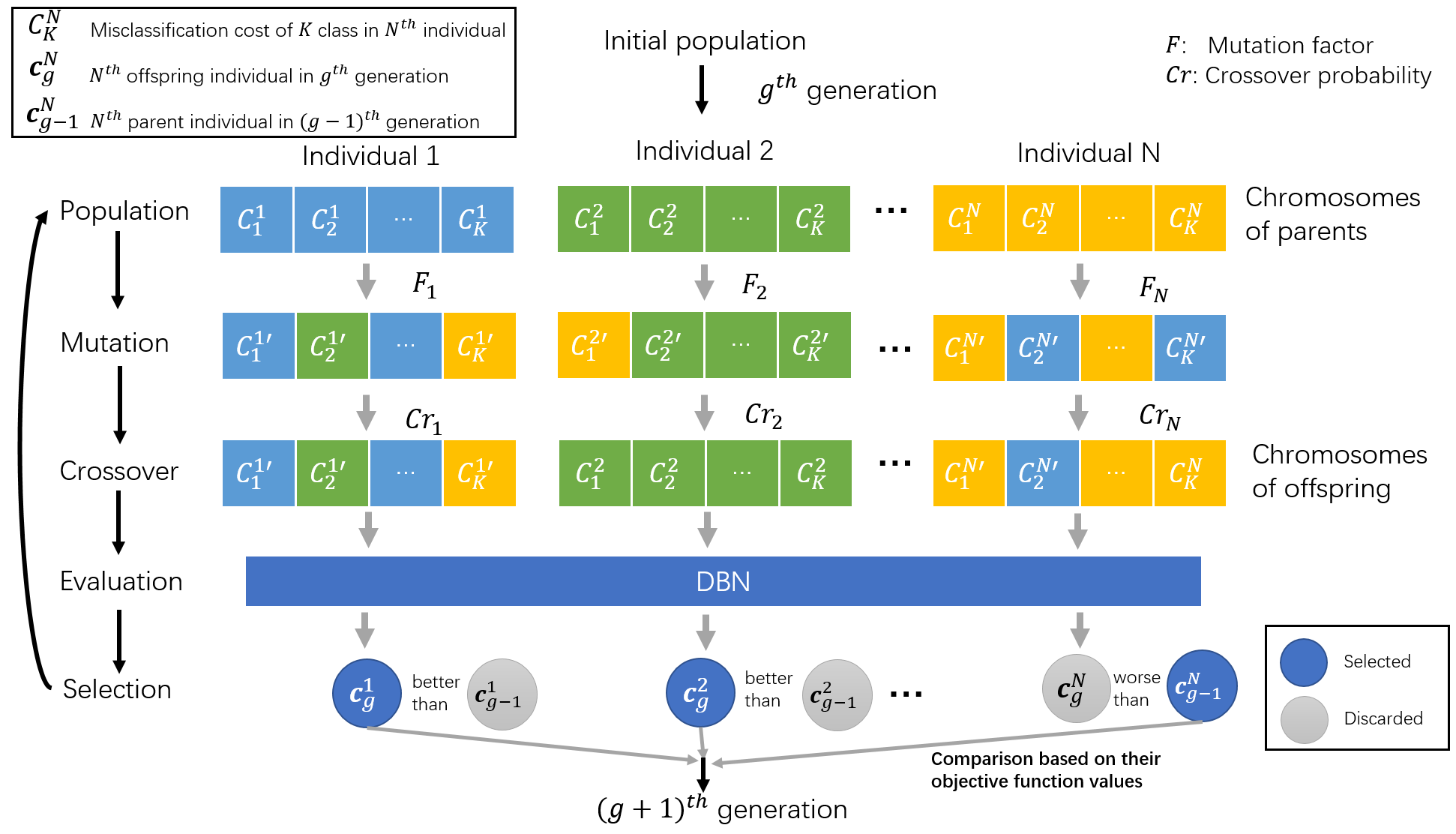}
\caption{ \revise{Illustration of chromosome encoding and evolution process in ECS-DBN. The chromosome is encoded with the misclassification costs of different classes in numerical type. The evolution process mainly includes mutation, crossover, evaluation and selection. The population is iteratively evolved via evolution process in each generation.}}

\label{fig:chromosome_evolution_ECS-DBN}
\end{figure}

Chromosome encoding is an important step in evolutionary algorithms which \rev{aims at effectively} representing important variables for better performance. In many real-world applications, misclassification costs in cost-sensitive deep belief network are usually unknown. \rev{In order to obtain appropriate costs, in our proposed approach }each chromosome represents misclassification costs for \rev{different classes}, and the final evolved best chromosome is chosen as the misclassification costs for \rev{ECS-DBN}. The chromosome encoding \rev{here} directly encodes the misclassification costs as values in the chromosome with numerical type and value range of $[0,1]$. \revise{Fig.~\ref{fig:chromosome_evolution_ECS-DBN} illustrates of chromosome encoding and evolution process in ECS-DBN.}

\subsection{Population Initialization}
The initial population is obtained via uniformly random sampling in feasible solution space for each variable within the specified range of the corresponding variable. 
\rev{The population is to hold possible misclassification costs and forms the unit of evolution. The evolution of the misclassification costs is an iterative process with the population in each iteration called a generation.}

\subsection{Adaptive DE Operators}
After initialization, adaptive differential evolution evolves the population with a sequence of three evolutionary operations, i.e. mutation, crossover, and selection, generation by generation. Mutation is carried out with DE mutation strategy to create mutation individuals based on the current parent population as shown in Step 2.1 of Algorithm~\ref{alg:ECS-DBN}. After mutation, a binomial crossover operation is utilized to generate the final offspring as shown in Step 2.2 of Algorithm~\ref{alg:ECS-DBN}. In adaptive DE, each individual has its associated crossover probability instead of a fixed value. The selection operation selects the best one from the parent individuals and offspring individuals according to their corresponding fitness values as shown in Step 2.3 of Algorithm~\ref{alg:ECS-DBN}. Parameter adaptation is conducted at each generation\revise{. In this way, the control parameters are automatically updated} to appropriate values \rev{without the need of prior parameter setting knowledge in DE}. The crossover probability of each individual is generated independently based on a normal distribution with mean $\mu_{Cr}$ and standard deviation 0.1. Similarly, the mutation factor of each individual is generated \rev{independently} based on a Cauchy distribution with location parameter $\mu_F$ and scale parameter 0.1. Both the mean $\mu_{Cr}$ and the location parameter $\mu_{F}$ are updated at the end of each generation as shown in Step 2.4 of Algorithm~\ref{alg:ECS-DBN}.

\subsection{Fitness Evaluation}
\rev{Fitness evaluation allows us to choose the appropriate misclassification costs.} In the proposed method, each individual chromosome is introduced into individual DBN as misclassification costs. \rev{We} generate suitable misclassification costs for DBN \rev{using the} training set. G-mean of training set is chosen as the objective \rev{function for the optimization.}

\subsection{Termination Condition}
Evolutionary algorithms are designed to evolve the population generation by generation and \revise{maintain the} convergence as well as diversity characteristics within the population. A maximum number of generation\rev{s} is set to be \revise{a} termination condition of the algorithm. \rev{In this implementation, we consider the solutions converged when the best fitness value remains unchanged over the past 30 generations~\cite{lim2016evolutionary}.} \revise{The algorithm terminates either when it reaches the maximum number of generations or when it meets the convergence condition. } 

\subsection{ECS-DBN Creation}
Eventually, the optimization process ends with the best individual which is used as misclassification costs to form an ECS-DBN. \revise{The best individual is obtained from the last generation.}

\section{Evaluation On Benchmark Datasets}\label{sec:benchmark_simulation}
In this section, \rev{the proposed \revise{ECS-DBN approach} is evaluated} on 58 popular KEEL benchmark datasets.

\subsection{Nomenclature}\label{sec:nomenclature}
ADASYN, SMOTE and its various resampling methods are applied with DBN to generate synthetic minority data on the imbalanced datasets. The nomenclature convention used in labeling the imbalance learning methods are as follows: the prefix letters ``ADASYN", ``SMOTE", ``SMOTE-SVM", ``SMOTE-borderline1", and ``SMOTE-borderline2" respectively, \rev{represent the} adaptive synthetic sampling approach~\cite{he2008adasyn}, Synthetic Minority Over-sampling Technique~\cite{chawla2002smote}, Support Vectors SMOTE~\cite{nguyen2011borderline}, and Borderline SMOTE of types 1 and 2~\cite{han2005borderline}. The suffix ``-DBN" represents deep belief network.

\subsection{Benchmark Datasets}
In this paper, benchmark datasets are selected from KEEL (Knowledge Extraction based on Evolutionary Learning) dataset repository~\cite{alcala2009keel}. The details specification of 58 binary-class imbalanced datasets are shown in Table~\ref{tab:dataset_spec}. All datasets are downloaded from KEEL website\footnote{http://sci2s.ugr.es/keel/imbalanced.php}. \revise{They are known to have a high imbalance ratio} between \revise{the} \rev{majority and minority} \revise{classes}. 

The imbalance ratio \rev{(IR)} is \revise{the number of data samples in majority class divided by that in minority class which is} \rev{described} by (\ref{eq:ir}).
\begin{equation}
\label{eq:ir}
\scriptsize IR = \frac{\#~majority}{\#~minority}
\end{equation}

\begin{table}[t]
\begin{center}
\caption{ \rev{A summary of the} KEEL binary-class imbalanced benchmark datasets. \rev{The details of each dataset includes its imbalance ratio, number of attributes, number of training data and number of test data. A total of 58 datasets used in this paper are listed below.}}
\label{tab:dataset_spec}
\begin{scriptsize}
\begin{tabular}{l|cccc}
\hline
\hline
Datasets                                                                    & \begin{tabular}[c]{@{}c@{}}Imbalance\\   Ratio(IR)\end{tabular} & Attributes & TrainData & TestData \\ \hline
abalone9-18                                                                 & 16.40 & 8  & 585  & 146  \\
cleveland-0-vs-4                                                          & 12.62 & 13 & 142  & 35   \\
ecoli-0-1-3-7-vs-2-6                                                      & 39.14 & 7  & 225  & 56   \\
ecoli-0-1-4-6-vs-5                                                        & 13.00 & 6  & 224  & 56   \\
ecoli-0-1-4-7-vs-2-3-5-6                                                  & 10.59 & 7  & 269  & 67   \\
ecoli-0-1-4-7-vs-5-6                                                      & 12.28 & 6  & 266  & 66   \\
ecoli-0-1-vs-2-3-5                                                        & 9.17  & 7  & 195  & 49   \\
ecoli-0-1-vs-5                                                            & 11.00 & 6  & 192  & 48   \\
ecoli-0-2-3-4-vs-5                                                        & 9.10  & 7  & 162  & 40   \\
ecoli-0-2-6-7-vs-3-5                                                      & 9.18  & 7  & 179  & 45   \\
ecoli-0-3-4-6-vs-5                                                        & 9.25  & 7  & 164  & 41   \\
ecoli-0-3-4-7-vs-5-6                                                      & 9.28  & 7  & 206  & 51   \\
ecoli-0-3-4-vs-5                                                          & 9.00  & 7  & 160  & 40   \\
ecoli-0-4-6-vs-5                                                          & 9.15  & 6  & 162  & 41   \\
ecoli-0-6-7-vs-3-5                                                        & 9.09  & 7  & 178  & 44   \\
ecoli-0-6-7-vs-5                                                          & 10.00 & 6  & 176  & 44   \\
ecoli3                                                                      & 8.6   & 7  & 269  & 67   \\
glass-0-1-2-3-vs-4-5-6                                                    & 3.2   & 9  & 171  & 43   \\
glass-0-1-4-6-vs-2                                                        & 11.06 & 9  & 164  & 41   \\
glass-0-1-5-vs-2                                                          & 9.12  & 9  & 138  & 34   \\
glass-0-1-6-vs-2                                                          & 10.29 & 9  & 153  & 39   \\
glass-0-1-6-vs-5                                                          & 19.44 & 9  & 147  & 37   \\
glass-0-4-vs-5                                                            & 9.22  & 9  & 74   & 18   \\
glass-0-6-vs-5                                                            & 11.00 & 9  & 86   & 22   \\
glass0                                                                      & 2.06  & 9  & 171  & 43   \\
glass1                                                                      & 1.82  & 9  & 171  & 43   \\
glass2                                                                      & 11.59 & 9  & 171  & 43   \\
glass4                                                                      & 15.47 & 9  & 171  & 43   \\
glass5                                                                      & 22.78 & 9  & 171  & 43   \\
glass6                                                                      & 6.38  & 9  & 171  & 43   \\
haberman                                                                    & 2.78  & 3  & 245  & 61   \\
iris0                                                                       & 2     & 4  & 120  & 30   \\
\begin{tabular}[c]{@{}l@{}}led7digit-0-2-4-5-6-7\\ -8-9-vs-1\end{tabular} & 10.97 & 7  & 354  & 89   \\
new-thyroid1                                                                & 5.14  & 5  & 172  & 43   \\
newthyroid2                                                                 & 5.14  & 5  & 172  & 43   \\
page-blocks-1-3-vs-4                                                      & 15.86 & 10 & 378  & 94   \\
page-blocks0                                                                & 8.79  & 10 & 4378 & 1094 \\
pima                                                                        & 1.87  & 8  & 614  & 154  \\
segment0                                                                    & 6.02  & 19 & 1846 & 462  \\
shuttle-c0-vs-c4                                                            & 13.87 & 9  & 1463 & 366  \\
shuttle-c2-vs-c4                                                            & 20.50 & 9  & 103  & 26   \\
vehicle0                                                                    & 3.25  & 18 & 677  & 169  \\
vowel0                                                                      & 9.98  & 13 & 790  & 198  \\
wisconsin                                                                   & 1.86  & 9  & 546  & 137  \\
yeast-0-2-5-6-vs-3-7-8-9                                                  & 9.14  & 8  & 803  & 201  \\
yeast-0-2-5-7-9-vs-3-6-8                                                  & 9.14  & 8  & 803  & 201  \\
yeast-0-3-5-9-vs-7-8                                                      & 9.12  & 8  & 405  & 101  \\
yeast-0-5-6-7-9-vs-4                                                      & 9.35  & 8  & 422  & 106  \\
yeast-1-2-8-9-vs-7                                                        & 30.57 & 8  & 757  & 188  \\
yeast-1-4-5-8-vs-7                                                        & 22.1  & 8  & 554  & 139  \\
yeast-1-vs-7                                                              & 14.30 & 7  & 367  & 92   \\
yeast-2-vs-4                                                              & 9.08  & 8  & 411  & 103  \\
yeast-2-vs-8                                                              & 23.10 & 8  & 385  & 97   \\
yeast1                                                                      & 2.46  & 8  & 1187 & 297  \\
yeast3                                                                      & 8.1   & 8  & 1187 & 297  \\
yeast4                                                                      & 28.10 & 8  & 1187 & 297  \\
yeast5                                                                      & 32.73 & 8  & 1187 & 297  \\
yeast6                                                                      & 41.40 & 8  & 1187 & 297      \\ \hline
\end{tabular}
\end{scriptsize}
\end{center}

\end{table}

\subsection{Implementation Details}
The learning rates of both pre-training and fine-tuning are 0.01. The number of pre-training and fine-tuning iterations are 100 and 300 respectively. The range of hidden neuron number is $[5, 50]$. The hidden neuron number of the networks are randomly selected from the range of hidden neuron number. \revise{Generally speaking, there are two key parameters that affect DE process namely mutation factor $F$ and crossover probability $Cr$. A larger $F$ enables DE of better exploration ability. A smaller $F$ allows DE to have better exploitation ability. DE with better exploitation ability leads to better convergence. DE with better exploration ability avoids local optima better, but it may result in slower convergence. Crossover probability affects the diversity of populations. A larger $Cr$ enables DE of better exploitation ability while a smaller $Cr$ enables DE of better exploration ability. We set the parameters empirically~\cite{zhang2009jade} to ensure that DE generally converges.} All the codes of resampling methods for comparison in this paper are from~\cite{JMLR:v18:16-365} in Python and their corresponding parameters are set as default. All the simulation results are obtained with 5-fold cross validation over 10 trials. All of the simulations are done on an Intel Core i5 3.20GHz machine with 16 GB RAM and NVIDIA GeForce GTX 980.

\subsection{Evaluation Metrics}\label{sec:benchmark_simulation_metric}

\rev{As in}~\cite{he2009learning,lim2016evolutionary,he2008adasyn,chawla2002smote,han2005borderline,ghazikhani2012class,nguyen2011borderline}, accuracy, G-mean, F1-score, recall, precision are the most commonly used evaluation metrics. Considering an imbalance binary-class classification problem, let TP, FP, FN, TN represent true positive, false positive, false negative and true negative, respectively. To evaluate the performance of a classifier, \rev{it is common to use the} overall \rev{accuracy that is} formulated in (\ref{eq:acccuracy}). 
\begin{equation}
\label{eq:acccuracy}
Accuracy=\frac{TP+TN}{TP+TN+FP+FN}
\end{equation}

In this section, both accuracy and G-mean are used\rev{. We use} G-mean~(\ref{eq:gmean}) \rev{because it evaluates} the degree of inductive bias which considers both positive and negative accuracy. The higher G-mean values represent the classifier could \rev{achieve} better performance on \rev{both minority and majority} classes. G-mean is less sensitive to data distributions \rev{that \revise{is given} as follows,}
\begin{equation}
 \label{eq:gmean}
 G \text{-} mean = \sqrt{\frac{TP}{TP+FN} \times \frac{TN}{TN+FP}}
\end{equation}

\subsection{Results of ECS-DBN}
\begin{figure}[t]
\centering
\includegraphics[width=3.5in,height=2in,clip,trim={1.25in 0.0in 2.6in 0.40in}]{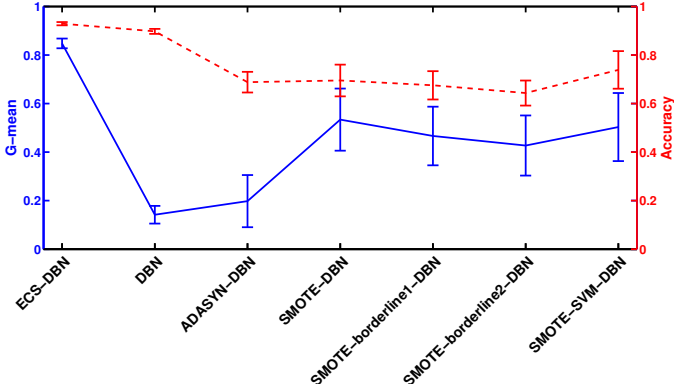}
\caption{ Comparison of the overall G-mean and accuracy \rev{across} 7 algorithms, i.e. ECS-DBN, DBN, ADASYN-DBN, SMOTE-DBN, SMOTE-borderline1-DBN, SMOTE-borderline2-DBN and SMOTE-SVM-DBN, on 58 benchmark datasets. \rev{ECS-DBN has higher average values and lower variance \revise{for} both G-mean and accuracy than other competing methods.}}
\label{fig:overall_benchmark_performance_ECS-DBN}

\end{figure}

In this section, \revise{we investigate the performance of DBN in different settings that include} ECS-DBN, DBN, ADASYN-DBN, SMOTE-DBN, SMOTE-borderline1-DBN, SMOTE-borderline2-DBN and SMOTE-SVM-DBN\revise{. We report the results over a} total \revise{of} runs on 58 KEEL benchmark datasets in terms of test accuracy and test G-mean, respectively. \revise{A detailed summary can be found at Tables~\ref{tab:imbalance_benchmark_result_accuracy} and~\ref{tab:imbalance_benchmark_result_gmean} in Annex, with the} best results \revise{being} highlighted in boldface. \rev{To visualize}, Fig.~\ref{fig:overall_benchmark_performance_ECS-DBN} illustrates the overall performances of \revise{the 7 variations of DBN algorithms} on 58 benchmark datasets. It is clear that ECS-DBN \revise{stands out from the rest}.

From the simulation results, the proposed ECS-DBN exhibits superior overall performance, especially in terms of G-mean. \revise{From Table~\ref{tab:imbalance_benchmark_result_gmean}, we observe} that ECS-DBN \rev{excels in 34} out of 58 benchmark datasets in terms of accuracy. \rev{As there are many more samples in majority class than minority class, a classifier can bias to the majority class yet achieve a high accuracy.} \rev{We also report results in G-mean, that} G-mean takes both performance of majority class and minority class into account. If some methods give \rev{a highly} biased performance, their G-mean values will be close to 0. It is \revise{worth noting} that ECS-DBN outperforms on 52 out of 58 benchmark datasets in terms of G-mean as shown in Table~\ref{tab:imbalance_benchmark_result_gmean}. \revise{We may attribute this to the fact that} ECS-DBN has been optimized using evolutionary algorithm with maximized G-mean objective. Therefore, the proposed ECS-DBN can provide better performances on minority class as well as those on majority class. 

\subsection{Computational Time Analysis}
\rev{The computing of \revise{ECS-DBN} at run-time is closely related to the DBN network complexity.} \rev{The} larger and deeper network size of DBN, \rev{the} \rev{more computing is required.} Table~\ref{tab:computational_time_benckmark_ECSDBN} reports the average computational time of ECS-DBN with 5-fold cross validation over 10 trials on the overall 58 benchmark datasets. In order to make \rev{a} fair and clear comparison between different imbalance learning methods, \rev{the} average computational time \rev{at run-time testing is summarized in} Table~\ref{tab:computational_time_benckmark_ECSDBN}. \rev{ECS-DBN shows a higher computational cost that is mainly due to the evolutionary algorithm.} \rev{We note that the} resampling methods are a little bit faster than ECS-DBN due to the small data size of KEEL benchmark datasets.

\begin{table}[t]

\caption{ Comparison of the average computational time \rev{across 7 algorithms} (i.e. ECS-DBN, DBN, ADASYN-DBN, SMOTE-DBN, SMOTE-borderline1-DBN, SMOTE-borderline2-DBN and SMOTE-SVM-DBN) with 5-fold cross validation on the overall 58 benchmark datasets over 10 trials.}
\label{tab:computational_time_benckmark_ECSDBN}
\centering
\resizebox{\columnwidth}{!}{
\begin{tabular}{l|cc}
\hline
Model Name            & \begin{tabular}[c]{@{}c@{}}Average\\ Computational\\ Time(s)\end{tabular} & \begin{tabular}[c]{@{}c@{}}Average Computational\\ Time without\\ DBN training time(s)\end{tabular} \\ \hline
ECS-DBN               & 26.64 $\pm$ 1.50                                                                & 18.23                                                                                               \\ 
ADASYN-DBN            & 22.97 $\pm$ 3.95                                                                & 14.56                                                                                               \\ 
SMOTE-SVM-DBN         & 24.03 $\pm$ 2.33                                                                & 15.62                                                                                               \\ 
SMOTE-borderline2-DBN & 24.53 $\pm$ 2.14                                                                & 16.12                                                                                               \\ 
SMOTE-borderline1-DBN & 25.52 $\pm$ 2.59                                                                & 17.11                                                                                               \\ 
SMOTE-DBN             & 25.81 $\pm$ 3.10                                                                & 17.40                                                                                               \\ 
DBN                   & 8.41 $\pm$ 1.28                                                                 & -                                                                                                   \\ \hline
\end{tabular}
}

\end{table}

\subsection{Statistical Tests for Evaluating Imbalance Learning}
\rev{Statistical tests provide evidence to ascertain the claim that the ECS-DBN outperforms other competitive methods.} \rev{It is noted that three common statistical tests }~\cite{garci2012evolutionary,galar2012review,lin2013dynamic,castro2013novel,wang2016online,lim2016evolutionary}, \rev{can \revise{serve} our purpose.} Wilcoxon paired signed-rank test \rev{is} adopted for pairwise comparisons between algorithms. Alternatively for comparison between $\xi$ multiple algorithms, a Holm post-hoc test can be utilized to conduct a $1 \times \xi$ posteriori tests between the control algorithm and the rest subgroups of algorithms. Average rank is also implemented for fair comparison.

\subsubsection{Wilcoxon Paired Signed-Rank Test}
In order to substantiate whether the results of ECS-DBN and other kinds of imbalance learning methods differ in a statistically significant way, a nonparametric statistical test known as Wilcoxon paired signed-rank test is conducted at the 5\% significance level. The Wilcoxon paired signed-rank test is employed separately between pairs of algorithms for each dataset. The entries which are significantly better than all the counterparts are marked with $\dag$ in Tables~\ref{tab:imbalance_benchmark_result_accuracy} and~\ref{tab:imbalance_benchmark_result_gmean}. The total number of win-lose-draw between the proposed method and its counterparts \rev{are} then reckoned. The pairwise comparisons of the proposed ECS-DBN method against other kinds of methods in terms of accuracy and G-mean are shown in Tables~\ref{tab:imbalance_benchmark_result_accuracy} and~\ref{tab:imbalance_benchmark_result_gmean}, respectively. In most cases, the proposed ECS-DBN method \revise{outperforms} other \rev{state-of-the-art} resampling methods, \rev{i.e. ADASYN, SMOTE, SMOTE-borderline1, SMOTE-borderline2 and SMOTE-SVM}. 

\subsubsection{Holm post-hoc Test}
For multiple comparisons, different algorithms \rev{are} compared using the Holm post-hoc test to detect statistical differences among them. The proposed ECS-DBN is chosen as the control algorithm for comparison. Then Holm post-hoc test \rev{is} implemented on the results of the method for all datasets in terms of accuracy and G-mean as shown in Tables~\ref{tab:imbalance_benchmark_result_accuracy} and~\ref{tab:imbalance_benchmark_result_gmean}, respectively. The $p$-values from Holm post-hoc test shown in Tables~\ref{tab:imbalance_benchmark_result_accuracy} and~\ref{tab:imbalance_benchmark_result_gmean} indicates \rev{that} the proposed ECS-DBN method statistically \rev{outperforms} other methods with significant statistical differences based on the results of all 58 datasets. \rev{In Holm \textit{post-hoc} test, ECS-DBN outperforms others.}

\subsubsection{Average Rank}
\begin{figure}[t]
\centering
\includegraphics[width=3.5in,,height=2in,clip,trim={1.42in 0.0in 2.7in 0.4in}]{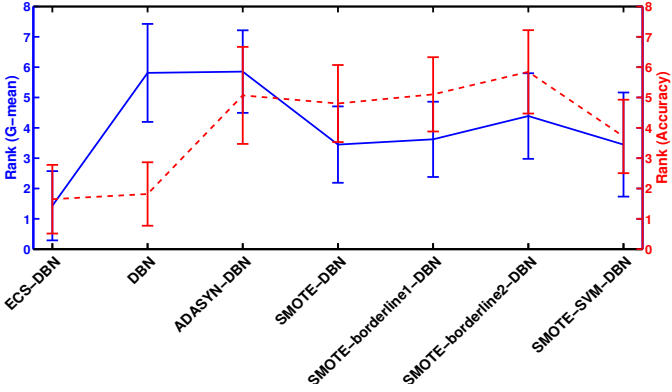}
\caption{ The average rank of 7 different algorithms, i.e. ECS-DBN, DBN, ADASYN-DBN, SMOTE-DBN, SMOTE-borderline1-DBN, SMOTE-borderline2-DBN and SMOTE-SVM-DBN, in terms of G-mean and accuracy on 58 KEEL benchmark datasets. \rev{ECS-DBN ranks first over other competing methods.}}
\label{fig:average_gmean_acc_ranking_benchmark_performance_ECS-DBN}

\end{figure}

Average rank is the mean of the ranks of individual method on all the datasets. Average ranks provide a fair comparison in terms of accuracy and G-mean of different methods as shown in Tables~\ref{tab:imbalance_benchmark_result_accuracy} and~\ref{tab:imbalance_benchmark_result_gmean}. Based on the average ranks in terms of both accuracy and G-mean on all datasets, the proposed ECS-DBN ranked first in the majority of the datasets. The results indicate \rev{that} ECS-DBN outperform\rev{s} other \rev{competing} methods and \rev{excels} especially in terms of G-mean. For \rev{a better} illustration, the average rank of different algorithms in terms of G-mean and accuracy is shown in Fig.~\ref{fig:average_gmean_acc_ranking_benchmark_performance_ECS-DBN}. \rev{It is apparent that ECS-DBN outranks others in terms of G-mean and accuracy.}
 
In sum, the results show that ECS-DBN method significantly outperforms \rev{other competing} methods. First of all, \rev{according to the} Wilcoxon paired signed-rank test, ECS-DBN outperform\rev{s} other methods in most cases. Secondly, the $p$-values from the Holm post-hoc test \revise{suggest that ECS-DBN achieves a statistically significant improvement over other competing methods.} Thirdly, average ranks show that ECS-DBN is ranked first \rev{across most of} the benchmark datasets. \rev{The fact that ECS-DBN outperforms DBN validates the need for cost-sensitive learning. Finally,} the significant improvement of ECS-DBN over some other methods \rev{manifests} the effectiveness of optimization. 

\section{Evaluation On A Real-World Dataset}\label{sec:real_world_experiment}
\subsection{Overview of the Imbalanced Gun Drilling Dataset}
\rev{We report the experiment results on gun drilling dataset collected from} a UNISIG USK25-2000 gun drilling machine in Advanced Manufacturing Lab \revise{at the} National University of Singapore, Singapore\rev{, in collaboration with} SIMTech-NUS joint lab. 

\subsection{Experimental Setup}
In the experiments, an Inconel 718 workpiece with the size of $1000mm*100mm*100mm$ is machined using gun drills. The tool diameter of gun drills is $8mm$. The detailed tool geometry of the tools are shown in Table~\ref{experimental_conditions}. Four vibration sensors (Kistler Type 8762A50) are mounted on the workpiece in order to measure the vibration signals in three directions (i.e. $x$, $y$ and $z$) during the gun drilling process. The details about sensor types and measurements are shown in Table~\ref{tab:sensors}. The sensor signals are acquired via a NI cDAQ-9178 data acquisition device and logged on a laptop.

In data acquisition, 14 channels of raw signals belonging to three types are logged. The measured signals include force signal, torque signal, and 12 vibration signals (i.e. acquired by 4 accelerometers in $x, y, z$ directions). The tool wears have been measured using Keyence VHX-5000 digital microscope. In this paper, the maximum flank wear \rev{which is most widely used in literature~\cite{zhang2016tool,zhang2017datadriven,Wang20141431,Biermann2015332,hong2017tool,xu2018gru}} has been used as the health indicator of the tool. In this dataset, it \rev{is} found 3 out of 20 tools are broken, 6 out of 20 tools have chipping at final state and 11 out of 20 tools are worn after gun drilling operations.

\begin{table}[t]
	\centering
	\caption{ \revise{A summary of the} 3 different types of sensors (i.e. accelerometer, dynamometer and microscope) used in the gun drilling experiments and the obtained measurements including vibration, force, torque and tool wear.}
	\label{tab:sensors}
	\begin{scriptsize}
	\resizebox{\columnwidth}{!}{
		\begin{tabular}{l|c|c|c}
			\hline
			Sensor type             & Vibration Sensor                                                                                                                       & \begin{tabular}[c]{@{}c@{}}Force\\Torque Sensor  \end{tabular}                                                             & Microscope                                                                     \\ \hline
			Description             & \begin{tabular}[c]{@{}c@{}c@{}}Kistler 50g 3-axis\\ accelerometer\\Type 8762A50 \end{tabular} & \begin{tabular}[c]{@{}c@{}c@{}c@{}}Dynamometer\\embedded in\\USK25-2000\\machine\end{tabular} & \begin{tabular}[c]{@{}c@{}c@{}c@{}}Keyence\\VHX-5000\\ Digital\\Microscope\end{tabular} \\ \hline
			\# Sensors              & 4                                                                                                                                      & 1                                                                                      & -                                                                              \\ \hline
			\# Channels per sensors & 3                                                                                                                                      & 2                                                                                      & -                                                                              \\ \hline
			Total number of channels       & 12                                                                                                                                     & 2                                                                                      & -                                                                              \\ \hline
			Measurements            & Vibration X,Y,Z                                                                                                                        & Thrust force, torque                                                                & Tool wear                                                                      \\ \hline
			Sampling frequency (Hz)          & 20,000                                                                                                                                  & 100                                                                                    & -                                                                              \\ \hline
		\end{tabular}
		}
		
	\end{scriptsize}
\end{table}

\begin{table}[!t]
	\centering
	\caption{\rev{Detailed number of data samples and imbalance ratio} of the imbalanced gun drilling dataset.}
	\label{tab:gun_drilling_dataset_description}
	\begin{tabular}{ll}
		\hline
		\hline
		Number of Signal Channels    & 14       \\ 
		Total Number of Data Samples & 19,712,414 \\ 
		Number of Training Samples   & 13,798,690 \\ 
		Number of Testing Samples    & 5,913,724  \\ 
		Imbalance Ratio & 10 \\\hline
	\end{tabular}
\end{table}

The machining operation is carried out with the detailed hole index, drill depth, tool geometry, tool diameter, feed rates, spindle speeds, machining times and tool final states are shown in Table~\ref{experimental_conditions}. The drilling depth is 50 mm in $z$-axis direction. The tool wear is captured and measured by Keyence digital microscope. The tool wear is measured after each drill during gun drilling operations.

\begin{table}[t]
	\centering
	\caption{\rev{Detailed real-world high aspect ratio deep hole gun drilling experimental conditions include tool diameter, spindle speed, feed rate and machining time for 20 drilling inserts with 6 different tool geometries.}}
	\label{experimental_conditions}
	\begin{scriptsize}
		\resizebox{\columnwidth}{!}{
			\begin{tabular}{c|c|c|c|c|c|c}
				\hline
				\begin{tabular}[c]{@{}c@{}}Hole\\Index\end{tabular} & \begin{tabular}[c]{@{}c@{}}Drill\\Depth\\(mm)\end{tabular} & Tool Geometry                   & \begin{tabular}[c]{@{}c@{}}Tool\\Diameter\\(mm)\end{tabular} & \begin{tabular}[c]{@{}c@{}}Spindle\\Speed\\(rpm)\end{tabular} & \begin{tabular}[c]{@{}c@{}}Feed\\Rate\\($\mu m$/rev)\end{tabular} & \begin{tabular}[c]{@{}c@{}}Machining\\Time(s)\end{tabular} \\ \hline
				H1\_01     & 50                                                          & \multirow{2}{*}{1450mm\_N4\_R9} & 8.05                                                          & 1200                                                           & 20                                                            & 125.00                                                                 \\ \cline{1-2} \cline{4-7} 
				H1\_02     & 50                                                          &                                 & 8.05                                                          & 1200                                                           & 20                                                            & 125.00                                                                   \\ \hline
				H2\_01     & 50                                                          & 1450mm\_N4\_R1                  & 8.05                                                          & 800                                                            & 20                                                            & 187.50                                                                 \\ \hline
				H2\_02     & 50                                                          & 1450mm\_N4\_R1                  & 8.05                                                          & 800                                                            & 20                                                            & 187.50                                                                   \\ \hline
				H2\_03     & 50                                                          & 1650mm\_N8\_R1                  & 8.02                                                          & 1650                                                           & 16                                                            & 113.64                                                                     \\ \hline
				H3\_01     & 50                                                          & \multirow{3}{*}{1450mm\_N8\_R9} & 8.05                                                          & 1650                                                           & 16                                                            & 113.64                                                                     \\ \cline{1-2} \cline{4-7} 
				H3\_02     & 50                                                          &                                 & 8.05                                                          & 1650                                                           & 16                                                            & 113.64                                                                   \\ \cline{1-2} \cline{4-7} 
				H3\_03     & 50                                                          &                                 & 8.05                                                          & 1650                                                           & 16                                                            & 113.64                                                                   \\ \hline
				H3\_04     & 50                                                          & \multirow{2}{*}{1450mm\_N8\_R9} & 8.05                                                          & 1650                                                           & 16                                                            & 113.64                                                             \\ \cline{1-2} \cline{4-7} 
				H3\_05     & 50                                                          &                                 & 8.05                                                          & 1650                                                           & 16                                                            & 113.64                                                              \\ \hline
				H3\_06     & 50                                                          & \multirow{2}{*}{1650mm\_N8\_R9} & 8.02                                                          & 1650                                                           & 16                                                            & 113.64                                                                \\ \cline{1-2} \cline{4-7} 
				H3\_07     & 50                                                          &                                 & 8.02                                                          & 1650                                                           & 16                                                            & 113.64                                                                \\ \hline
				H3\_08     & 50                                                          & \multirow{2}{*}{1650mm\_N8\_R9} & 8.02                                                          & 1650                                                           & 16                                                            & 113.64                                                                 \\ \cline{1-2} \cline{4-7} 
				H3\_09     & 50                                                          &                                 & 8.02                                                          & 1650                                                           & 16                                                            & 113.64                                                               \\ \hline
				H3\_10     & 50                                                          & 1650mm\_N8\_R1                  & 8.02                                                          & 1650                                                           & 16                                                            & 113.64                                                              \\ \hline
				H4\_01     & 50                                                          & \multirow{3}{*}{1219mm\_N8\_R9} & 8.08                                                          & 1650                                                           & 16                                                            & 113.64                                                                     \\ \cline{1-2} \cline{4-7} 
				H4\_02     & 50                                                          &                                 & 8.08                                                          & 1650                                                           & 16                                                            & 113.64                                                                 \\ \cline{1-2} \cline{4-7} 
				H4\_03     & 50                                                          &                                 & 8.08                                                          & 1650                                                           & 16                                                            & 113.64                                                                  \\ \hline
				H5\_01     & 50                                                          & 1450mm\_N8\_R9                  & 8.045                                                         & 1650                                                           & 16                                                            & 113.64                                                              \\ \hline
				H5\_02     & 50                                                          & 1450mm\_N8\_R9                  & 8.045                                                         & 1650                                                           & 16                                                            & 113.64                                                           \\ \hline
			\end{tabular}
		}
	\end{scriptsize}
\end{table}

Table~\ref{tab:gun_drilling_dataset_description} lists the details of the imbalanced gun drilling dataset. The imbalanced gun drilling dataset is selected from the raw experimental data by discarding lousy noise data samples. The total number of data samples in the imbalanced gun drilling dataset is 19,712,414. The number of training data samples and test data samples are 13,798,690 and 5,913,724 respectively. The data has been labeled into healthy (i.e. maximum flank wear of the tool $ < 300\mu m$) and faulty (i.e. maximum flank wear of the tool $ \geq 300\mu m$) two classes. The imbalance ratio (IR) of this dataset is 10. The data preprocessing and time window process are the same with~\cite{zhang2017datadriven}.

\begin{table*}[t]
	\centering
	\caption{\rev{Comparing} the performance of between Deep Belief Network (DBN) and Support Vector Machine (SVM), Multilayer Perceptron (MLP), K-Nearest Neighbor Classifier (KNN), Gradient Boosting (GB), Logistic Regression (LR), AdaBoost classifier, Lasso, and SGD different machine learning algorithms on gun drilling imbalanced dataset in terms of accuracy, G-Mean, AUC, precision, F1-score. \rev{DBN outperforms other competing methods in terms of all evaluation metrics.}}
	\label{tab:imbalance_binary_classification_results_without_cost_sensitive}
		\begin{scriptsize}
	\begin{tabular}{l|c|c|c|c|c}
		\hline
		Model Name                 & Accuracy               & G-mean                & AUC             & Precision              & F1-score             
		\\ \hline
		DBN                        & \textbf{0.9954 $\pm$ 0.0006} & \textbf{0.9830 $\pm$ 0.0027} & \textbf{0.9831 $\pm$ 0.0027} & \textbf{0.9968 $\pm$ 0.0005} & \textbf{0.9975 $\pm$ 0.0003} 
		\\ \hline
		SVM                        & 0.9894 $\pm$ 0.0103$^\dag$          & 0.8943 $\pm$ 0.3143          & 0.9443 $\pm$ 0.1562          & 0.9898 $\pm$ 0.0284          & 0.9944 $\pm$ 0.0148        
		\\ \hline
		MLP                        & 0.9683 $\pm$ 0.0064          & 0.8492 $\pm$ 0.0350$^\dag$          & 0.8601 $\pm$ 0.0296$^\dag$          & 0.9733 $\pm$ 0.0056          & 0.9827 $\pm$ 0.0035       
		\\ \hline
		KNN       & 0.9733 $\pm$ 0.0026          & 0.8460 $\pm$ 0.0214$^\dag$          & 0.8579 $\pm$ 0.0181$^\dag$          & 0.9725 $\pm$ 0.0035          & 0.9855 $\pm$ 0.0014$^\dag$        
		\\ \hline
		GB & 0.9821 $\pm$ 0.0353          & 0.8088 $\pm$ 0.3910$^\dag$          & 0.9025 $\pm$ 0.1947$^\dag$          & 0.9822 $\pm$ 0.0354          & 0.9906 $\pm$ 0.0185$^\dag$        
		\\ \hline
		LR         & 0.9412 $\pm$ 0.0096          & 0.5903 $\pm$ 0.1058$^\dag$          & 0.6791 $\pm$ 0.0545$^\dag$          & 0.9398 $\pm$ 0.0095          & 0.9687 $\pm$ 0.0049$^\dag$        
		\\ \hline
		AdaBoost         & 0.9361 $\pm$ 0.0126          & 0.5362 $\pm$ 0.1246$^\dag$          & 0.6510 $\pm$ 0.0719$^\dag$          & 0.9349 $\pm$ 0.0127          & 0.9661 $\pm$ 0.0065        
		\\ \hline
		Lasso                      & 0.9523 $\pm$ 0.0414          & 0.5065 $\pm$ 0.4817$^\dag$          & 0.7401 $\pm$ 0.2303$^\dag$          & 0.9524 $\pm$ 0.0416          & 0.9749 $\pm$ 0.0216$^\dag$        
		\\\hline
		SGD              & 0.9258 $\pm$ 0.0169$^\dag$          & 0.3860 $\pm$ 0.2692$^\dag$          & 0.6098 $\pm$ 0.0894$^\dag$          & 0.9281 $\pm$ 0.0162          & 0.9606 $\pm$ 0.0095$^\dag$        
		\\ \hline

	\end{tabular}\\
	$\dag$ indicates that the difference between marked algorithm and the proposed algorithm is statistically significant using Wilcoxon rank sum test at the $5\%$ significance level.
	\end{scriptsize}

\end{table*}

\begin{table*}[t]
	\centering
	\caption{\rev{Comparing} the performance of ECS-DBN, CSDBN, SMOTE-SVM-DBN, ADASYN-DBN, SMOTE-borderline1-DBN, SMOTE-borderline2-DBN, SMOTE-DBN, DBN, \rev{weighted extreme learning machine (WELM)} \revise{and ECO-ensemble} on gun drilling imbalanced dataset in terms of accuracy, G-mean, AUC, precision, F1-score. \rev{ECS-DBN shows better performance in terms of G-mean, AUC and precision.}}
	\label{tab:imbalance_binary_classification_results_with_cost_sensitive}
	\begin{scriptsize}
	\begin{tabular}{l|c|c|c|c|c}
		\hline
		Model Name            & Accuracy               & G-mean                & AUC             & Precision              & F1-score             
		\\ \hline
		ECS-DBN                & 0.9960 $\pm$ 0.0002          & \textbf{0.9946 $\pm$ 0.0011} & \textbf{0.9946 $\pm$ 0.0011} & \textbf{0.9996 $\pm$ 0.0002} & 0.9978 $\pm$ 0.0001   
		\\ \hline
		CSDBN                 & 0.9957 $\pm$ 0.0012          & 0.9916 $\pm$ 0.0026          & 0.9916 $\pm$ 0.0025          & 0.9987 $\pm$ 0.0007          & 0.9977 $\pm$ 0.0006          
		\\ \hline
		SMOTE-SVM-DBN         & 0.9961 $\pm$ 0.0003          & 0.9858 $\pm$ 0.0011          & 0.9859 $\pm$ 0.0011          & 0.9974 $\pm$ 0.0002$^\dag$         & 0.9978 $\pm$ 0.0002          
		\\ \hline
		ADASYN-DBN            & \textbf{0.9964 $\pm$ 0.0004} & 0.9857 $\pm$ 0.0016          & 0.9858 $\pm$ 0.0016$^\dag$          & 0.9973 $\pm$ 0.0003          & \textbf{0.9980 $\pm$ 0.0002} 
		\\ \hline
		SMOTE-borderline2-DBN & 0.9962 $\pm$ 0.0001          & 0.9852 $\pm$ 0.0010$^\dag$          & 0.9853 $\pm$ 0.0010          & 0.9972 $\pm$ 0.0002          & 0.9979 $\pm$ 0.0001          
		\\ \hline
		SMOTE-borderline1-DBN & 0.9958 $\pm$ 0.0003          & 0.9837 $\pm$ 0.0021          & 0.9838 $\pm$ 0.0020          & 0.9969 $\pm$ 0.0004          & 0.9977 $\pm$ 0.0002          
		\\ \hline
		SMOTE-DBN             & 0.9959 $\pm$ 0.0001          & 0.9837 $\pm$ 0.0010$^\dag$          & 0.9838 $\pm$ 0.0010          & 0.9969 $\pm$ 0.0002$^\dag$          & 0.9978 $\pm$ 0.0001$^\dag$          
		\\ \hline
		DBN                   & 0.9954 $\pm$ 0.0006          & 0.9830 $\pm$ 0.0027          & 0.9831 $\pm$ 0.0027$^\dag$          & 0.9968 $\pm$ 0.0005          & 0.9963 $\pm$ 0.0003$^\dag$          
		\\ \hline
		WELM & 0.7397 $\pm$ 0.0177$^\dag$ & 0.7532 $\pm$ 0.0170$^\dag$ & 0.7841 $\pm$ 0.0159$^\dag$ & 0.6274 $\pm$ 0.0202$^\dag$ & 0.7261 $\pm$ 0.0148$^\dag$ \\ \hline
		\revise{ECO-ensemble} & \revise{0.9895 $\pm$ 0.0014$^\dag$} & \revise{0.9713 $\pm$ 0.0133$^\dag$} & \revise{0.9694 $\pm$ 0.0185$^\dag$} & \revise{0.9836 $\pm$ 0.0105$^\dag$} & \revise{0.9925 $\pm$ 0.0042$^\dag$} \\ \hline		
	\end{tabular}	\\
	$\dag$ indicates that the difference between marked algorithm and the proposed algorithm is statistically significant using Wilcoxon rank sum test at the $5\%$ significance level.
	\end{scriptsize}\\

\end{table*}

\begin{figure}[t]
\centering
\includegraphics[width=3.5in,clip,trim={1in 0.3in 1.9in 0.1in}]{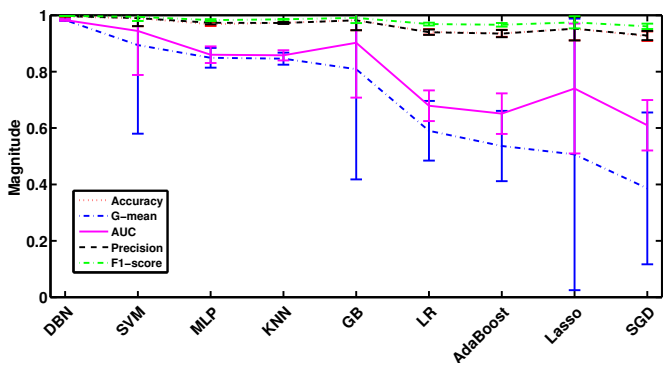}
\caption{ Illustration \rev{of} the performance between different machine learning algorithms, i.e. DBN and SVM, MLP, KNN, GB, LR, AdaBoost, Lasso, and SGD, on gun drilling imbalanced dataset in terms of accuracy, G-Mean, AUC, precision, F1-score. \rev{DBN provides higher average values on all evaluation metrics than other methods.}}
\label{fig:cmp_gun_drillinig_performance_DBN_others}
\end{figure}

\begin{figure}[t]
\centering
\includegraphics[width=3.5in,clip,trim={0.7in 0.63in 3.1in 0.1in}]{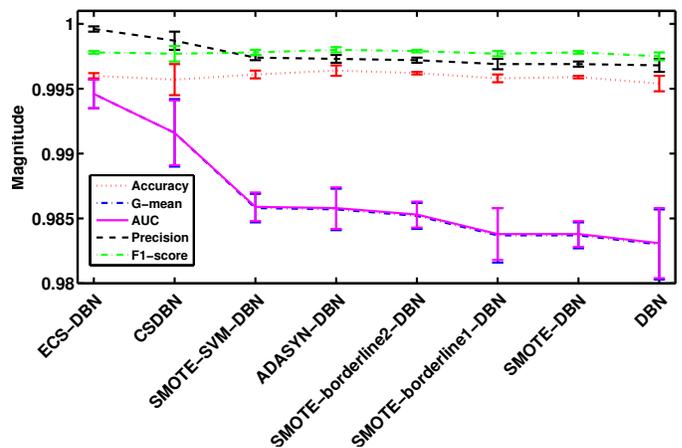}
\caption{ Illustration the performance of between different imbalance learning methods, i.e. ECS-DBN, CSDBN, SMOTE-SVM-DBN, ADASYN-DBN, SMOTE-borderline1-DBN, SMOTE-borderline2-DBN, SMOTE-DBN and baseline DBN, on gun drilling imbalanced dataset in terms of accuracy, G-Mean, AUC, precision, F1-score. \rev{ECS-DBN shows better performance than other resampling methods in terms of G-mean, AUC and precision.}}
\label{fig:cmp_gun_drillinig_performance_ECS-DBN}

\end{figure}

The details of the gun drilling cycle are as follows.
\begin{enumerate}
	\item Machine startup.
	\item Feed internal coolant through coolant hole of gun drill.
	\item Start to drill through the workpiece.
	\item Finish drilling and pull the tool back.
	\item Machine shutdown.
\end{enumerate}
The internally-fed coolant will exhaust the heat generated during gun drilling process and give high accuracy and precision performance.

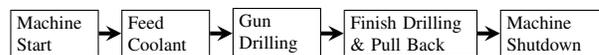
\begin{figure}[!t]
	\centering
	\begin{tikzpicture}[>=stealth, auto, node distance=3cm]
	\begin{scriptsize}
	
	\node[draw,shape=rectangle,text width=1cm] (a) at (0,0) {Machine Start};
	\node[draw,shape=rectangle,fill=white,right=0.3 of a,text width=1cm] (b) {Feed Coolant};
	\node[draw,shape=rectangle,fill=white,right=0.3 of b,text width=1cm] (c) {Gun Drilling};
	\node[draw,shape=rectangle,fill=white,right=0.3 of c,text width=1.6cm] (d) {Finish Drilling \& Pull Back};
	\node[draw,shape=rectangle,fill=white,right=0.3 of d,text width=1.2cm] (e) {Machine Shutdown};
	
	\draw[line width=1.5pt,->] (a) -- (b);
	\draw[line width=1.5pt,->] (b) -- (c);
	\draw[line width=1.5pt,->] (c) -- (d);
	\draw[line width=1.5pt,->] (d) -- (e);
	\end{scriptsize}
	\end{tikzpicture}
	\caption{The procedure of gun drilling experiments includes 5 steps, namely start machine, feed coolant, start gun grilling in the workpiece, finish drilling \& pull back the drill and finally shutdown the machine.}
	\label{fig:gun_drilling_procedure}
\end{figure}

\subsection{Evaluation Metrics}
In this section, despite the evaluation metrics used in Section~\ref{sec:benchmark_simulation_metric}, AUC, precision and F1-Score are introduced to evaluate the methods. The formulation of those metrics are listed below:
\begin{equation}
\label{eq:precision}
 Precision=\frac{TP}{TP+FP}
\end{equation}
\begin{equation}
\label{eq:recall}
 Recall=\frac{TP}{TP+FN}
\end{equation}
\begin{equation}
\label{eq:f1score}
 F1\textendash score=2 \cdot \frac{precision \times recall}{precision+recall}
\end{equation}
Precision~(\ref{eq:precision}) is a measure of a classifiers exactness. For this real-world application, exactness of classifier is an important indicator. F1-score~(\ref{eq:f1score}) is a weighted average of precision and recall. The reason for \rev{choosing} F1-score in this real-world application is that F1-score is used to evaluate the performance of the minority class (i.e. faulty) which is very important in this application. G-mean and F1-score incorporate both to express their tradeoff~\cite{he2009learning} and indicate the overall performance. AUC is used to evaluate the overall performance of the method on both classes. Recall is also known as \rev{the} true positive rate\rev{,} which signifies a measure of completeness. 

\subsection{\rev{Experiment} Results}
In this section, \rev{all the parameters of DBN and} adaptive DE are the same \revise{with those in} Section~\ref{sec:benchmark_simulation}. All conventional machine learning algorithms for comparison purpose in this paper are from~\cite{scikit-learn} and their corresponding parameters are set as default. The resampling methods for comparison in this section are the same with Section~\ref{sec:benchmark_simulation}. Since there is only one real-world dataset, only Wilcoxon paired signed-rank test has been implemented in this section. 

The simulation results of imbalanced gun drilling dataset with DBN, multi-layer neural network (MLP), support vector machine (SVM), K-nearest neighbors (KNN), linear classifier with stochastic gradient descent (SGD) training, logistic regression (LR), gradient boosting (GB), AdaBoost classifier, Lasso are shown in Table~\ref{tab:imbalance_binary_classification_results_without_cost_sensitive} in terms of classification accuracy, G-means, Precision and F1-score \rev{on test data}. For better illustration, Fig.~\ref{fig:cmp_gun_drillinig_performance_DBN_others} presents the errorbar plot of the performance between different algorithms evaluated with different metrics. All the simulation results include the average performances and the corresponding standard deviation values. The simulation results are obtained on test data. Based on the simulation results, it is obvious that DBN outperforms other 9 conventional machine learning algorithms. This could contribute to the strong feature learning ability of DBN. Therefore, with its automatically hierarchical feature learning ability, DBN is chosen as the base classifier for this real-world application. 

Table~\ref{tab:imbalance_binary_classification_results_with_cost_sensitive} presents the comparison between ECS-DBN, CSDBN, \rev{weighted extreme learning machine (WELM)}, \revise{ECO-ensemble} and various resampling methods including ADASYN, SMOTE, SMOTE-borderline1, SMOTE-borderline2 and SMOTE-SVM in terms of accuracy, G-mean, Precision, F1-score. For clear illustration, Fig.~\ref{fig:cmp_gun_drillinig_performance_ECS-DBN} shows the errorbar plot comparison of the performance between ECS-DBN and different resampling methods with different evaluation metrics. \rev{WELM~\cite{Zong2013weighted} is a state-of-the-art cost-sensitive extreme learning machine.} \revise{ECO-ensemble~\cite{lim2016evolutionary} incorporates synthetic data generation within an ensemble framework optimized by EA simultaneously.} The experiment results show that ECS-DBN outperforms WELM \revise{and ECO-ensemble}. By comparing with other resampling methods, ECS-DBN outperforms on G-mean and precision metrics. Especially on G-mean, ECS-DBN generates a significant performance improvement \revise{over} the others. 

\begin{figure}[t]
\centering
\begin{subfigure}[b]{0.45\linewidth}
    \includegraphics[width=1.7in,clip,trim={0.5in 0.1in 0.4in 0.1in}]{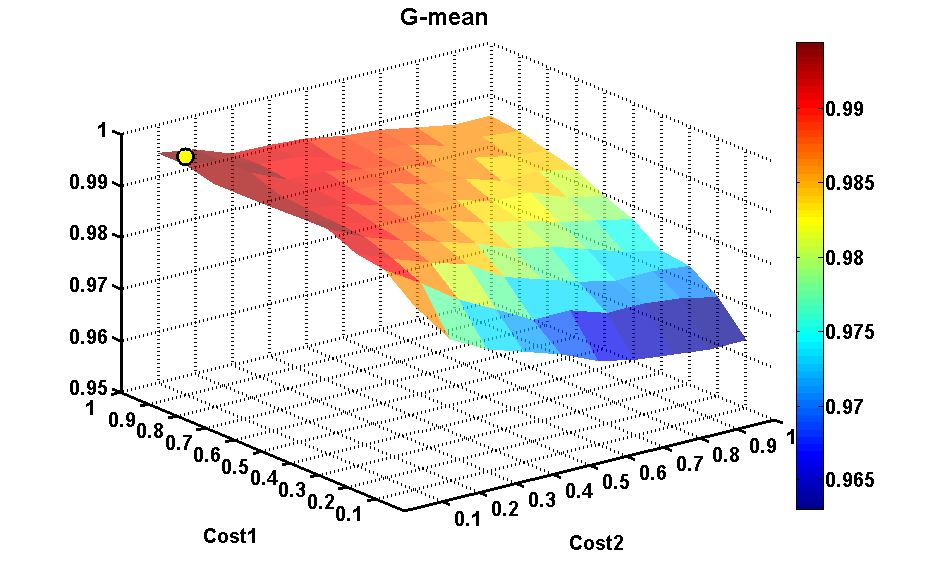}
    \caption{G-mean}
    \label{fig:cmp_gun_drillinig_performance_ECSDBN_gmean_gridsearch}
\end{subfigure}
\hspace{0.01\linewidth}
\begin{subfigure}[b]{0.45\linewidth}
\includegraphics[width=1.7in,clip,trim={0.4in 0.1in 0.4in 0.1in}]{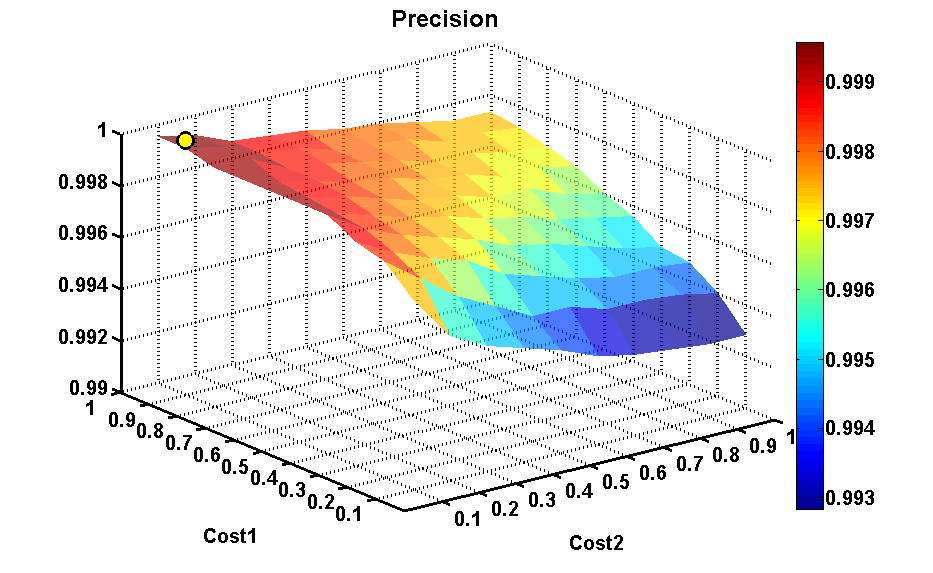}
    \caption{Precision}
    \label{fig:cmp_gun_drillinig_performance_ECSDBN_precision_gridsearch}
\end{subfigure}
\caption{ \rev{Illustration of the comparison between the performance of the proposed ECS-DBN and grid search of misclassification costs on DBN in terms of G-mean and Precision. Note that the solid yellow circle represents the performance of ECS-DBN. This example indicates the proposed ECS-DBN can obtain better performance by finding suitable misclassification costs.}}
\label{fig:cmp_gun_drillinig_performance_ECSDBN_gridsearch}

\end{figure}

\revise{As an example, we illustrate the G-mean and precision between ECS-DBN and the grid search of misclassification costs of DBN in Fig.~\ref{fig:cmp_gun_drillinig_performance_ECSDBN_gridsearch}. It is clear that the proposed ECS-DBN benefits from the well optimized misclassification costs to achieve better performance.} In terms of accuracy and F1-score, ECS-DBN can also provide comparable performance. The performance improvement of ECS-DBN over DBN and CSDBN with randomly generated cost values on many performance metrics further illustrates the need for cost-sensitive learning and the effectiveness of optimization. Therefore, ECS-DBN could generate comparable performance not only on benchmark dataset but also on real-world application. 

\begin{figure}[t]
\centering
\begin{subfigure}[b]{0.45\linewidth}
    \includegraphics[width=1.7in,clip,trim={0.4in 0.1in 0.4in 0.1in}]{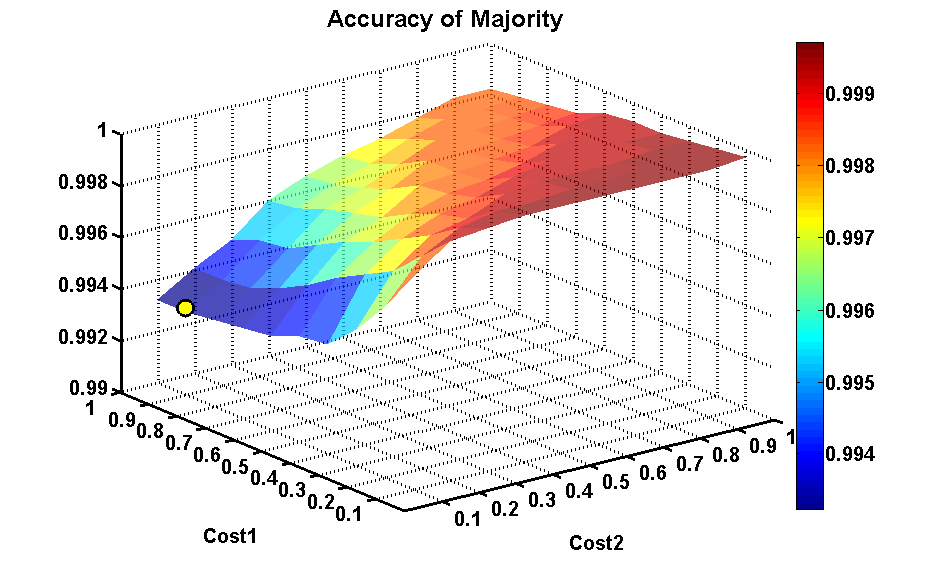}
    \caption{Accuracy of majority \rev{class}}
    \label{fig:cmp_gun_drillinig_performance_ECSDBN_recall_gridsearch}
\end{subfigure}
\hspace{0.01\linewidth}
\begin{subfigure}[b]{0.45\linewidth}
\includegraphics[width=1.7in,clip,trim={0.4in 0.1in 0.5in 0.1in}]{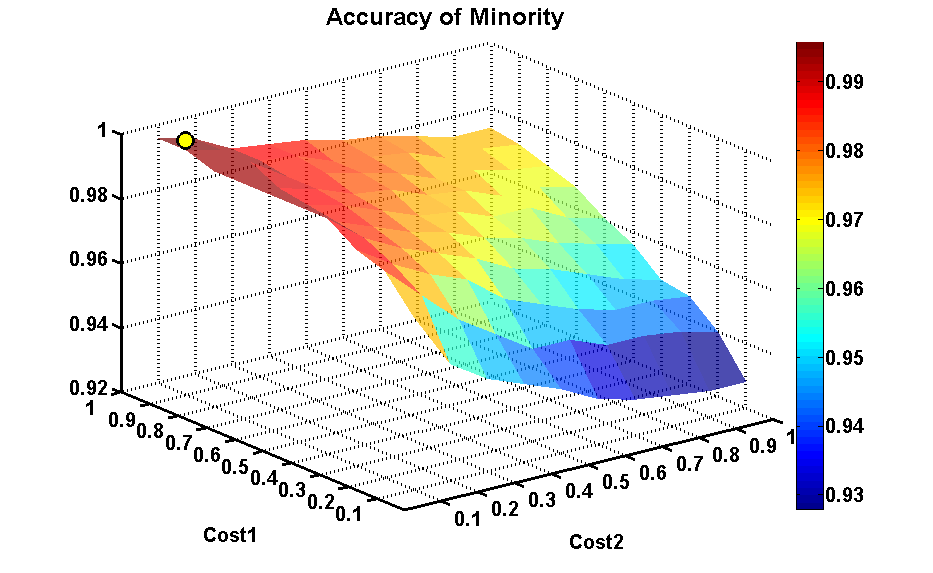}
    \caption{Accuracy of minority \rev{class}}
    \label{fig:cmp_gun_drillinig_performance_ECSDBN_acc-_gridsearch}
\end{subfigure}
\caption{ \rev{Illustration of the comparison between the accuracy of majority class and minority class respectively with the proposed ECS-DBN and grid search of misclassification costs on DBN. Note that the solid yellow circle denotes the performance of ECS-DBN.}}
\label{fig:cmp_gun_drillinig_performance_ECSDBN_acc_gridsearch}

\end{figure}

\revise{We further examine the effect of the proposed ECS-DBN over majority verse minority classes.} \revise{Fig.~\ref{fig:cmp_gun_drillinig_performance_ECSDBN_acc_gridsearch} shows that ECS-DBN benefits from more suitable misclassification costs and improves} the accuracy of minority class. Fig.~\ref{fig:cmp_gun_drillinig_performance_ECSDBN_gridsearch} and Fig.~\ref{fig:cmp_gun_drillinig_performance_ECSDBN_acc_gridsearch} \revise{validate the ability of ECS-DBN of finding} suitable misclassification costs via evolutionary algorithm \revise{that improves} the accuracy of minority class\rev{, thus the overall performance.} \revise{How ECS-DBN may impact on the majority class depends on the way we define the objective functions. While ECS-DBN improves the overall performance, it also provides a mechanism to trade off the performance between the majority class and the minority class.}

\subsection{Computational \rev{Cost}}
Average computational time of ECS-DBN, DBN, ADASYN-DBN, SMOTE-DBN, SMOTE-borderline1-DBN, SMOTE-borderline2-DBN and SMOTE-SVM-DBN on the gun drilling imbalanced dataset are presented in Table~\ref{tab:computational_time_gundrilling_ECSDBN}. It is obvious that ECS-DBN \revise{consumes less} average computational time than other \rev{competing} methods. \revise{In comparison with the} KEEL benchmark datasets, the gun drilling imbalanced dataset has \revise{a} much larger size of data samples which increases the computational complexity for resampling methods. Hence, the proposed ECS-DBN is more efficient than some resampling methods to large dataset. \revise{If we compare the computational time required between the evolutionary algorithm to estimate the misclassification cost, and the DBN training, the former is very small and negligible. In short, the proposed ECS-DBN approach is both efficient and effective.}

\begin{table}[t]
\centering
\caption{ Comparison of the average computational time \rev{across 7 algorithms} (i.e. ECS-DBN, DBN, ADASYN-DBN, SMOTE-DBN, SMOTE-borderline1-DBN, SMOTE-borderline2-DBN and SMOTE-SVM-DBN) with 5-fold cross validation on the gun drilling imbalanced dataset over 10 trials.}
\label{tab:computational_time_gundrilling_ECSDBN}
\begin{scriptsize}
\resizebox{\columnwidth}{!}{
\begin{tabular}{l|cc}
\hline
Model Name            & \begin{tabular}[c]{@{}c@{}}Average\\ Computational\\ Time(s)\end{tabular} & \begin{tabular}[c]{@{}c@{}}Average Computational\\ Time without\\ DBN training time(s)\end{tabular} \\ \hline
ECS-DBN               & 9977.39 $\pm$ 148.55                                                            & \textbf{1280.28}                                                                                    \\ 
ADASYN-DBN            & 12697.79 $\pm$ 1287.39                                                          & 4000.68                                                                                             \\ 
SMOTE-DBN             & 13992.46 $\pm$ 1772.37                                                          & 5295.35                                                                                             \\ 
SMOTE-borderline1-DBN & 14031.35 $\pm$ 1135.55                                                          & 5334.24                                                                                             \\ 
SMOTE-borderline2-DBN & 14310.19 $\pm$ 764.11                                                           & 5613.08                                                                                             \\ 
SMOTE-SVM-DBN         & 20942.94 $\pm$ 6609.65                                                          & 12245.83                                                                                            \\ 
DBN                   & 8697.11 $\pm$ 8308.22                                                          & -                                                                                                   \\ \hline
\end{tabular}
}
\end{scriptsize}

\end{table}

\section{Conclusion}\label{sec:conclusion}
In this paper, an evolutionary cost-sensitive deep belief network (ECS-DBN) \rev{is proposed} for \rev{imbalanced} classification problem. \revise{We have shown that} ECS-DBN \revise{significantly outperforms} other competing techniques on 58 benchmark datasets and a real-world dataset. The proposed ECS-DBN improves DBN by \rev{applying} cost-sensitive learning strategy. To tackle with unknown misclassification costs in practice, adaptive differential evolution algorithm has been utilized to find \rev{the} misclassification costs. Since many real-world data are \rev{naturally} imbalanced\revise{, therefore,} the \rev{misclassification} costs of different classes are \revise{usually} unknown, ECS-DBN \rev{offers an effective solution.} \revise{ECS-DBN is also computationally more efficient} than some \rev{popular} resampling methods on large \rev{scale} datasets. \revise{It} can also be \rev{easily implemented on} multi-class \revise{scenarios}. In this paper, \revise{we only incorporate the} cost-sensitive learning technique on algorithmic level. However, the imbalanced distribution in feature space may also impact the performance of learning models. In \revise{the future, we consider that} cost-sensitive methods could also be applied \revise{to high dimensional data and dynamic data}. Furthermore, online learning usually suffers from concept drift with different imbalance ratio over time. \revise{ECS-DBN} can be further extended for online imbalanced classification problems \rev{with some online learning strategies}. \revise{The proposed approach can also be applied to other deep learning models such as convolutional neural network, etc.} 

\section*{Acknowledgment}
\revi{Chong Zhang and Haizhou Li were supported by Neuromorphic Computing Program, RIE2020 AME Programmatic Grant, A*STAR, Singapore.}

\ifCLASSOPTIONcaptionsoff
  \newpage
\fi
\bibliographystyle{IEEEtran}
\begin{scriptsize}
\bibliography{Database,imbalance_learning,tcm,publications_zc_abbr}
\end{scriptsize}


\begin{IEEEbiography}[{\includegraphics[width=1in,height=1.25in,clip,keepaspectratio]{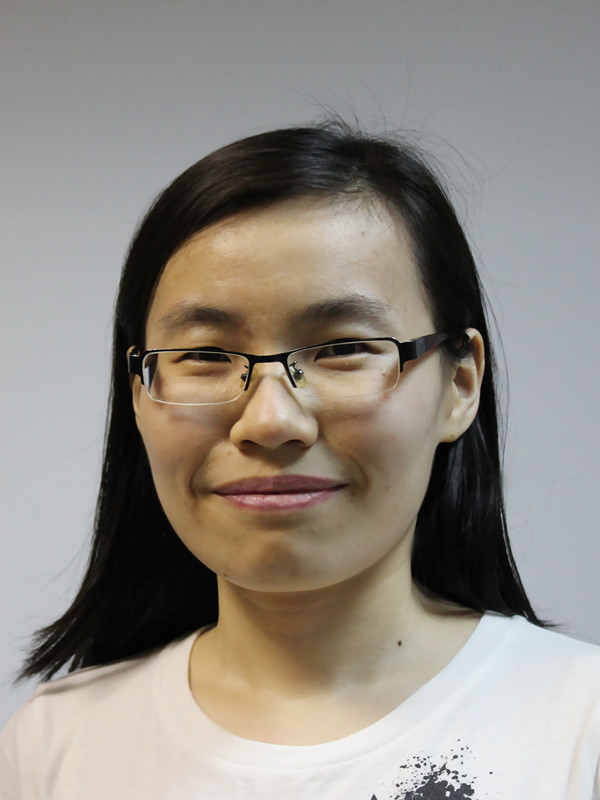}}]{Chong~Zhang} received the B.Eng. degree in Engineering from Harbin Institute of Technology, China, and the M.Sc. degree from National University of Singapore in 2011 and 2012, respectively. She is currently a Ph.D. student as well as a research engineer in Department of Electrical and Computer Engineering at National University of Singapore. 
	
Her research interests include computational intelligence, machine learning/deep learning, data science and their applications in big data analytics, diagnostics, prognostics, health condition monitoring, voice conversion, etc. 
\end{IEEEbiography}

\begin{IEEEbiography}[{\includegraphics[width=1in,height=1.25in,clip,keepaspectratio]{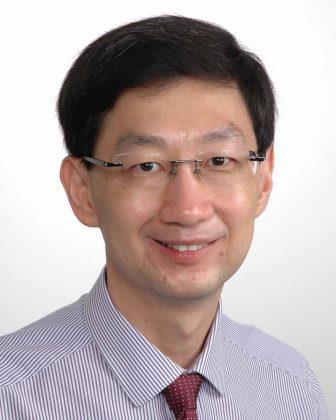}}]{Kay~Chen~Tan}
(SM'08--F'14) 
is a full Professor with the Department of Computer Science, City University of Hong Kong. He is the Editor-in-Chief of IEEE Transactions on Evolutionary Computation, was the EiC of IEEE Computational Intelligence Magazine (2010-2013), and currently serves on the Editorial Board member of 20+ journals. He is an elected member of IEEE CIS AdCom (2017-2019). He has published 200+ refereed articles and 6 books. He is a Fellow of IEEE.

\end{IEEEbiography}
\begin{IEEEbiography}[{\includegraphics[width=1in,height=1.25in,clip,keepaspectratio]{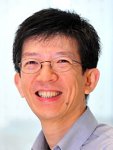}}]{Haizhou~Li} 
(M'91--SM'01--F'14) received the B.Sc., M.Sc., and Ph.D degree in electrical and electronic engineering from South China University of Technology, Guangzhou, China in 1984, 1987, and 1990 respectively. Dr Li is currently a Professor at the Department of Electrical and Computer Engineering, National University of Singapore (NUS). He is also a Conjoint Professor at the University of New South Wales, Australia. His research interests include automatic speech recognition, speaker/language recognition, and natural language processing.

Prior to joining NUS, he taught in the University of Hong Kong (1988-1990) and South China University of Technology (1990-1994). He was a Visiting Professor at CRIN in France (1994-1995), Research Manager at the Apple-ISS Research Centre (1996-1998), Research Director in Lernout \& Hauspie Asia Pacific (1999-2001), Vice President in InfoTalk Corp. Ltd. (2001-2003), and the Principal Scientist and Department Head of Human Language Technology in the Institute for Infocomm Research, Singapore (2003-2016).
Dr Li is currently the Editor-in-Chief of IEEE/ACM Transactions on Audio, Speech and Language Processing (2015-2018), a Member of the Editorial Board of Computer Speech and Language (2012-2018). He was an elected Member of IEEE Speech and Language Processing Technical Committee (2013-2015), the President of the International Speech Communication Association (2015-2017), the President of Asia Pacific Signal and Information Processing Association (2015-2016), and the President of Asian Federation of Natural Language Processing (2017-2018). He was the General Chair of ACL 2012 and INTERSPEECH 2014. Dr Li is a Fellow of the IEEE. 

He was a recipient of the National Infocomm Award 2002 and the President’s Technology Award 2013 in Singapore. He was named one of the two Nokia Visiting Professors in 2009 by the Nokia Foundation. 
\end{IEEEbiography}

\begin{IEEEbiography}[{\includegraphics[width=1in,height=1.25in,clip,keepaspectratio]{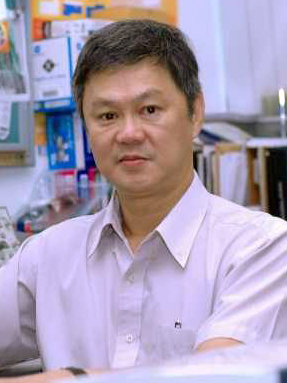}}]{Geok~Soon~Hong}
received the B.Eng. degree in Control Engineering in 1982 from University of Sheffield, UK. He was awarded a university scholarship to further his studies and obtained a Ph.D. degree in control engineering in 1987. The topic of his research was in stability and performance analysis for systems with multi-rate sampling problems. He is now an Associate Professor at the Department of Mechanical Engineering, National University of Singapore (NUS), Singapore. His research interests are in Control Theory, Multirate sampled data system, Neural network Applications and Industrial Automation, Modeling and control of Mechanical Systems, Tool Condition Monitoring, AI techniques in monitoring and Diagnostics.
\end{IEEEbiography}

\newpage
\beginsupplement

\section*{Annex}

\begin{table*}[!h]
	\centering
	\caption{Summary of the test accuracy across 7 different algorithms, i.e. ECS-DBN, DBN and a group of resampling methods including ADASYN-DBN, SMOTE-DBN, SMOTE-borderline1-DBN, SMOTE-borderline2-DBN, SMOTE-SVM-DBN, on 58 KEEL benchmark datasets. ECS-DBN outperforms DBN and a group of resampling methods on 34 out of 58 benchmark datasets.}
	\begin{scriptsize}
		\label{tab:imbalance_benchmark_result_accuracy}
		\begin{tabular}{l|ccccccc}
			\hline
			\hline
			\multirow{4}{*}{Datasets}      &  \multirow{4}{*}{ECS-DBN} & \multirow{4}{*}{DBN} & \multicolumn{5}{c}{Resampling Methods}\\\cline{4-8}\\
			& &                    & ADASYN-DBN      & SMOTE-DBN     & \begin{tabular}[c]{@{}c@{}}SMOTE -borderline1\\-DBN \end{tabular} & \begin{tabular}[c]{@{}c@{}}SMOTE-borderline2\\-DBN \end{tabular} & \begin{tabular}[c]{@{}c@{}}SMOTE-SVM\\-DBN \end{tabular} \\ \hline
			abalone9-18                      & 0.9307$\pm$0.0048                 & \textbf{0.9403$\pm$0.0045} & 0.7674$\pm$0.0061 & 0.6510$\pm$0.0457 & 0.7217$\pm$0.0754 & 0.7014$\pm$0.0954 & 0.7033$\pm$0.0177          \\
			cleveland-0\_vs\_4               & \textbf{0.9858$\pm$0.0115}        & 0.9467$\pm$0.0115          & 0.6939$\pm$0.1419 & 0.7585$\pm$0.2254 & 0.8317$\pm$0.1145 & 0.7915$\pm$0.1242 & 0.8293$\pm$0.1698          \\
			ecoli-0-1-3-7\_vs\_2-6           & 0.9697$\pm$0.0001                 & \textbf{0.9859$\pm$0.0052} & 0.9031$\pm$0.0966 & 0.8482$\pm$0.1337 & 0.8438$\pm$0.1397 & 0.7810$\pm$0.1407 & 0.9188$\pm$0.0744          \\
			ecoli-0-1-4-6\_vs\_5             & 0.9582$\pm$0.0001                 & \textbf{0.9729$\pm$0.0146} & 0.7273$\pm$0.0000 & 0.8454$\pm$0.1201 & 0.6092$\pm$0.1756 & 0.7354$\pm$0.1552 & 0.8500$\pm$0.1328          \\
			ecoli-0-1-4-7\_vs\_2-3-5-6       & 0.9512$\pm$0.0005                 & \textbf{0.9647$\pm$0.0000} & 0.6148$\pm$0.0518 & 0.7260$\pm$0.1890 & 0.8909$\pm$0.1020 & 0.7253$\pm$0.1524 & 0.7188$\pm$0.2138          \\
			ecoli-0-1-4-7\_vs\_5-6           & \textbf{0.9622$\pm$0.0002}        & 0.9560$\pm$0.0218          & 0.6661$\pm$0.1150 & 0.7773$\pm$0.1822 & 0.7929$\pm$0.1830 & 0.8714$\pm$0.0674 & 0.8019$\pm$0.1890          \\
			ecoli-0-1\_vs\_2-3-5             & \textbf{0.9750$\pm$0.0042}        & 0.9689$\pm$0.0000          & 0.7143$\pm$0.0502 & 0.6491$\pm$0.1826 & 0.7209$\pm$0.1478 & 0.6209$\pm$0.1444 & 0.6782$\pm$0.2011          \\
			ecoli-0-1\_vs\_5                 & 0.9498$\pm$0.0049                 & \textbf{0.9686$\pm$0.0058} & 0.7200$\pm$0.0000 & 0.8318$\pm$0.1031 & 0.8091$\pm$0.1695 & 0.7236$\pm$0.1463 & 0.8900$\pm$0.1042          \\
			ecoli-0-2-3-4\_vs\_5             & \textbf{0.9730$\pm$0.0051$^\dag$} & 0.9029$\pm$0.0101          & 0.6613$\pm$0.0000 & 0.7253$\pm$0.2016 & 0.5527$\pm$0.1755 & 0.5560$\pm$0.1098 & 0.7582$\pm$0.1888          \\
			ecoli-0-2-6-7\_vs\_3-5           & 0.9475$\pm$0.0067                 & \textbf{0.9772$\pm$0.0186} & 0.6049$\pm$0.0000 & 0.5040$\pm$0.1659 & 0.3851$\pm$0.0189 & 0.3832$\pm$0.0219 & 0.5505$\pm$0.2114          \\
			ecoli-0-3-4-6\_vs\_5             & 0.9441$\pm$0.0009                 & \textbf{0.9596$\pm$0.0284} & 0.6719$\pm$0.0000 & 0.6860$\pm$0.2410 & 0.6774$\pm$0.1987 & 0.4785$\pm$0.0913 & 0.6860$\pm$0.2461          \\
			ecoli-0-3-4-7\_vs\_5-6           & 0.9358$\pm$0.0079                 & \textbf{0.9477$\pm$0.0212} & 0.6158$\pm$0.0832 & 0.6526$\pm$0.1581 & 0.7155$\pm$0.2012 & 0.6534$\pm$0.1423 & 0.7250$\pm$0.1821          \\
			ecoli-0-3-4\_vs\_5               & \textbf{0.9480$\pm$0.0094}        & 0.9007$\pm$0.0079          & 0.6774$\pm$0.0000 & 0.7344$\pm$0.1932 & 0.7767$\pm$0.1818 & 0.5689$\pm$0.1526 & 0.8133$\pm$0.1103          \\
			ecoli-0-4-6\_vs\_5               & \textbf{0.9741$\pm$0.0253$^\dag$} & 0.9431$\pm$0.0326          & 0.6984$\pm$0.0000 & 0.6641$\pm$0.1879 & 0.6391$\pm$0.1857 & 0.5174$\pm$0.1388 & 0.7391$\pm$0.1852          \\
			ecoli-0-6-7\_vs\_3-5             & 0.9564$\pm$0.0100                 & \textbf{0.9696$\pm$0.0121} & 0.6000$\pm$0.0651 & 0.6300$\pm$0.2025 & 0.5010$\pm$0.1641 & 0.4720$\pm$0.1193 & 0.7270$\pm$0.1862          \\
			ecoli-0-6-7\_vs\_5               & 0.8664$\pm$0.0077                 & \textbf{0.9636$\pm$0.0000} & 0.6184$\pm$0.0000 & 0.5980$\pm$0.1729 & 0.5470$\pm$0.1443 & 0.4980$\pm$0.1305 & 0.5730$\pm$0.1939          \\
			ecoli3                           & \textbf{0.9567$\pm$0.0050$^\dag$} & 0.8936$\pm$0.0082          & 0.5920$\pm$0.0000 & 0.7212$\pm$0.1748 & 0.7245$\pm$0.1734 & 0.7033$\pm$0.1344 & 0.8252$\pm$0.1220          \\
			glass-0-1-2-3\_vs\_4-5-6         & 0.9597$\pm$0.0074                 & \textbf{0.9815$\pm$0.0136} & 0.5795$\pm$0.1187 & 0.7415$\pm$0.1704 & 0.5402$\pm$0.1153 & 0.4829$\pm$0.0427 & 0.7768$\pm$0.2115          \\
			glass-0-1-4-6\_vs\_2             & \textbf{0.9890$\pm$0.0038$^\dag$} & 0.9038$\pm$0.0000          & 0.7286$\pm$0.0000 & 0.4915$\pm$0.0580 & 0.4872$\pm$0.0551 & 0.4777$\pm$0.0498 & 0.6667$\pm$0.0000          \\
			glass-0-1-5\_vs\_2               & \textbf{0.9356$\pm$0.0214$^\dag$} & 0.8864$\pm$0.0000          & 0.6066$\pm$0.0000 & 0.5154$\pm$0.0426 & 0.5128$\pm$0.0472 & 0.4974$\pm$0.0132 & 0.7193$\pm$0.0000          \\
			glass-0-1-6\_vs\_2               & 0.6711$\pm$0.0002                 & \textbf{0.8980$\pm$0.0000} & 0.6667$\pm$0.0000 & 0.4795$\pm$0.0407 & 0.5273$\pm$0.0398 & 0.5170$\pm$0.0497 & 0.6377$\pm$0.0000          \\
			glass-0-1-6\_vs\_5               & 0.6761$\pm$0.0110                 & \textbf{0.9364$\pm$0.0000} & 0.5325$\pm$0.0551 & 0.6432$\pm$0.2084 & 0.6261$\pm$0.2060 & 0.7011$\pm$0.1824 & 0.7627$\pm$0.0000          \\
			glass-0-4\_vs\_5                 & \textbf{0.9346$\pm$0.0001$^\dag$} & 0.8763$\pm$0.0067          & 0.6000$\pm$0.0935 & 0.5524$\pm$0.1713 & 0.5286$\pm$0.1475 & 0.5000$\pm$0.0502 & 0.5738$\pm$0.2204          \\
			glass-0-6\_vs\_5                 & \textbf{0.9636$\pm$0.0288$^\dag$} & 0.8941$\pm$0.0329          & 0.6409$\pm$0.0144 & 0.5000$\pm$0.0422 & 0.5760$\pm$0.1632 & 0.5600$\pm$0.1323 & 0.6180$\pm$0.1153          \\
			glass0                           & \textbf{0.9513$\pm$0.0175$^\dag$} & 0.6664$\pm$0.0250          & 0.5231$\pm$0.0000 & 0.5542$\pm$0.0963 & 0.5347$\pm$0.0672 & 0.4681$\pm$0.0262 & 0.5597$\pm$0.0980          \\
			glass1                           & \textbf{0.7196$\pm$0.0282$^\dag$} & 0.6487$\pm$0.0527          & 0.4444$\pm$0.0000 & 0.5145$\pm$0.0765 & 0.4986$\pm$0.0730 & 0.5101$\pm$0.0658 & 0.5246$\pm$0.0482          \\
			glass2                           & \textbf{0.9721$\pm$0.0001$^\dag$} & 0.9091$\pm$0.0000          & 0.6216$\pm$0.0000 & 0.4485$\pm$0.0630 & 0.4768$\pm$0.0916 & 0.4424$\pm$0.0506 & 0.7538$\pm$0.0000          \\
			glass4                           & \textbf{0.9717$\pm$0.0088$^\dag$} & 0.9276$\pm$0.0103          & 0.6024$\pm$0.0151 & 0.6851$\pm$0.2207 & 0.6733$\pm$0.2186 & 0.6495$\pm$0.2120 & 0.7667$\pm$0.0205          \\
			glass5                           & \textbf{0.9680$\pm$0.0002}        & 0.9455$\pm$0.0088          & 0.6000$\pm$0.0000 & 0.6058$\pm$0.1948 & 0.6330$\pm$0.1653 & 0.6495$\pm$0.1836 & 0.7671$\pm$0.0000          \\
			glass6                           & \textbf{0.9732$\pm$0.0002$^\dag$} & 0.8613$\pm$0.0077          & 0.6821$\pm$0.0622 & 0.7570$\pm$0.1686 & 0.7118$\pm$0.2084 & 0.7237$\pm$0.1157 & 0.7656$\pm$0.1745          \\
			haberman                         & \textbf{0.9821$\pm$0.0001$^\dag$} & 0.7301$\pm$0.0041          & 0.6162$\pm$0.0000 & 0.4655$\pm$0.0385 & 0.4478$\pm$0.0140 & 0.5035$\pm$0.0450 & 0.4841$\pm$0.0336          \\
			iris0                            & \textbf{0.9363$\pm$0.0050}        & 0.7227$\pm$0.0000          & 0.6842$\pm$0.0000 & 0.9340$\pm$0.0608 & 0.7158$\pm$0.0999 & 0.7447$\pm$0.1278 & 0.6256$\pm$0.1913          \\
			\begin{tabular}[c]{@{}l@{}}led7digit-0-2-4-5-6-7\\ -8-9\_vs\_1\end{tabular} & 0.7385$\pm$0.0063                 & \textbf{0.9141$\pm$0.0075} & 0.8211$\pm$0.0075 & 0.8722$\pm$0.0070 & 0.8877$\pm$0.1123 & 0.8291$\pm$0.0925 & 0.9034$\pm$0.0048          \\
			new-thyroid1                     & \textbf{0.9681$\pm$0.0026$^\dag$} & 0.8451$\pm$0.0000          & 0.4815$\pm$0.0000 & 0.6244$\pm$0.1967 & 0.4867$\pm$0.0729 & 0.4844$\pm$0.0420 & 0.4911$\pm$0.0491          \\
			newthyroid2                      & \textbf{0.9870$\pm$0.0130$^\dag$} & 0.8460$\pm$0.0000          & 0.4815$\pm$0.0000 & 0.5478$\pm$0.1106 & 0.5011$\pm$0.1051 & 0.4733$\pm$0.0375 & 0.5100$\pm$0.0686          \\
			page-blocks-1-3\_vs\_4           & \textbf{0.9796$\pm$0.0089$^\dag$} & 0.9419$\pm$0.0011          & 0.7617$\pm$0.0673 & 0.6860$\pm$0.1413 & 0.8414$\pm$0.0687 & 0.7874$\pm$0.1129 & 0.7450$\pm$0.0213          \\
			page-blocks0                     & 0.8639$\pm$0.0028                 & \textbf{0.9048$\pm$0.0000} & 0.8312$\pm$0.0129 & 0.8597$\pm$0.0149 & 0.8059$\pm$0.0151 & 0.6984$\pm$0.1074 & 0.8499$\pm$0.0292          \\
			pima                             & \textbf{0.9630$\pm$0.0105$^\dag$} & 0.6555$\pm$0.0192          & 0.5766$\pm$0.0288 & 0.6188$\pm$0.0637 & 0.5608$\pm$0.0629 & 0.5232$\pm$0.0651 & 0.5184$\pm$0.0618          \\
			segment0                         & \textbf{0.9546$\pm$0.0054}        & 0.8630$\pm$0.0007          & 0.9197$\pm$0.0167 & 0.8235$\pm$0.1277 & 0.8776$\pm$0.1010 & 0.8055$\pm$0.1055 & 0.9385$\pm$0.0132          \\
			shuttle-c0-vs-c4                 & 0.9175$\pm$0.0001                 & \textbf{0.9342$\pm$0.0000} & 0.9389$\pm$0.0000 & 0.8971$\pm$0.0170 & 0.9391$\pm$0.0007 & 0.9393$\pm$0.0009 & 0.8998$\pm$0.0988          \\
			shuttle-c2-vs-c4                 & \textbf{0.9701$\pm$0.0050}        & 0.9395$\pm$0.0214          & 0.7180$\pm$0.0569 & 0.6339$\pm$0.1474 & 0.6032$\pm$0.2077 & 0.6355$\pm$0.1427 & 0.6875$\pm$0.0000          \\
			vehicle0                         & \textbf{0.8968$\pm$0.0054$^\dag$} & 0.7740$\pm$0.0066          & 0.7272$\pm$0.0644 & 0.7577$\pm$0.1420 & 0.7497$\pm$0.1447 & 0.6735$\pm$0.1276 & 0.7420$\pm$0.1501          \\
			vowel0                           & \textbf{0.9311$\pm$0.0029}        & 0.9089$\pm$0.0114          & 0.8935$\pm$0.0694 & 0.9053$\pm$0.0207 & 0.8323$\pm$0.1675 & 0.8472$\pm$0.0184 & 0.8976$\pm$0.0448          \\
			wisconsin                        & 0.9716$\pm$0.0060                 & 0.6910$\pm$0.0054          & 0.9683$\pm$0.0026 & 0.9216$\pm$0.0482 & 0.9266$\pm$0.0510 & 0.9203$\pm$0.0777 & \textbf{0.9743$\pm$0.0149} \\
			yeast-0-2-5-6\_vs\_3-7-8-9       & \textbf{0.9298$\pm$0.0033}        & 0.9019$\pm$0.0028          & 0.6954$\pm$0.0787 & 0.7576$\pm$0.1378 & 0.7404$\pm$0.0982 & 0.7775$\pm$0.0075 & 0.8201$\pm$0.0891          \\
			yeast-0-2-5-7-9\_vs\_3-6-8       & \textbf{0.9577$\pm$0.0074}        & 0.9047$\pm$0.0073          & 0.7416$\pm$0.1150 & 0.7945$\pm$0.1842 & 0.6722$\pm$0.1298 & 0.6667$\pm$0.1163 & 0.8285$\pm$0.1103          \\
			yeast-0-3-5-9\_vs\_7-8           & \textbf{0.9916$\pm$0.0083$^\dag$} & 0.8976$\pm$0.0041          & 0.6725$\pm$0.0000 & 0.6136$\pm$0.0577 & 0.5952$\pm$0.1207 & 0.5649$\pm$0.0980 & 0.6333$\pm$0.0516          \\
			yeast-0-5-6-7-9\_vs\_4           & 0.7561$\pm$0.0075                 & \textbf{0.9030$\pm$0.0055} & 0.6674$\pm$0.0118 & 0.8121$\pm$0.0260 & 0.8280$\pm$0.0499 & 0.7285$\pm$0.1245 & 0.8385$\pm$0.1318          \\
			yeast-1-2-8-9\_vs\_7             & 0.9279$\pm$0.0028                 & \textbf{0.9664$\pm$0.0103} & 0.7098$\pm$0.0000 & 0.6614$\pm$0.1196 & 0.6963$\pm$0.0784 & 0.6229$\pm$0.0987 & 0.7111$\pm$0.0000          \\
			yeast-1-4-5-8\_vs\_7             & \textbf{0.9661$\pm$0.0001}        & 0.9540$\pm$0.0082          & 0.7308$\pm$0.0000 & 0.5422$\pm$0.0833 & 0.6440$\pm$0.0705 & 0.5542$\pm$0.0676 & 0.7325$\pm$0.0000          \\
			yeast-1\_vs\_7                   & \textbf{0.9715$\pm$0.0134$^\dag$} & 0.9371$\pm$0.0013          & 0.6325$\pm$0.0000 & 0.6884$\pm$0.0385 & 0.7163$\pm$0.0534 & 0.6670$\pm$0.0699 & 0.7958$\pm$0.1101          \\
			yeast-2\_vs\_4                   & \textbf{0.9825$\pm$0.0037$^\dag$} & 0.9026$\pm$0.0000          & 0.6684$\pm$0.0054 & 0.8681$\pm$0.0299 & 0.6457$\pm$0.1348 & 0.6159$\pm$0.0910 & 0.8095$\pm$0.1702          \\
			yeast-2\_vs\_8                   & 0.9598$\pm$0.0001                 & \textbf{0.9917$\pm$0.0033} & 0.8243$\pm$0.0154 & 0.6883$\pm$0.1186 & 0.5320$\pm$0.0734 & 0.5255$\pm$0.0911 & 0.6662$\pm$0.0786          \\
			yeast1                           & 0.7122$\pm$0.0107                 & \textbf{0.7125$\pm$0.0000} & 0.5518$\pm$0.0107 & 0.5278$\pm$0.1033 & 0.5072$\pm$0.0544 & 0.4867$\pm$0.0474 & 0.6106$\pm$0.0916          \\
			yeast3                           & 0.8906$\pm$0.0034                 & \textbf{0.8911$\pm$0.0042} & 0.6316$\pm$0.0000 & 0.7808$\pm$0.1711 & 0.6965$\pm$0.1725 & 0.6321$\pm$0.1174 & 0.6378$\pm$0.1809          \\
			yeast4                           & \textbf{0.9651$\pm$0.0023}        & 0.9651$\pm$0.0028          & 0.8048$\pm$0.1011 & 0.8230$\pm$0.1177 & 0.8813$\pm$0.1098 & 0.8996$\pm$0.0027 & 0.9291$\pm$0.0032          \\
			yeast5                           & 0.9704$\pm$0.0023                 & \textbf{0.9728$\pm$0.0020} & 0.9606$\pm$0.0008 & 0.8715$\pm$0.0976 & 0.8703$\pm$0.0948 & 0.9272$\pm$0.0013 & 0.9238$\pm$0.0489          \\
			yeast6                           & 0.9759$\pm$0.0042                 & \textbf{0.9825$\pm$0.0019} & 0.8215$\pm$0.1405 & 0.8477$\pm$0.1229 & 0.8932$\pm$0.0389 & 0.8190$\pm$0.1656 & 0.9534$\pm$0.0019         
			\\
			\hline
			Average Rank & \textbf{1.6466} & 1.8190 & 5.0690 & 4.8017  & 5.1034 & 5.8448 & 3.7155 \\ 
			Win-lose-draw& - & 34-4-20 & 50-0-8 & 48-0-10 & 48-1-9 &	54-0-4    & 44-0-14    \\
			Holm post-hoc Test & - & \textbf{3.60922E-03} & \textbf{3.44739E-11}
			& \textbf{5.76082E-11} & \textbf{6.06238E-11}		  & \textbf{3.10875E-11}		   & \textbf{2.12982E-10}  \\
			\hline      
		\end{tabular}
		\\
		$\dag$ indicates that the difference between the proposed algorithm and all other compared algorithm is statistically significant using pair-wised Wilcoxon rank sum test at the $5\%$ significance level.
	\end{scriptsize}
\end{table*}

\begin{table*}[!h]
	\centering
	\caption{Summary of test G-mean results across 7 different algorithms, i.e. ECS-DBN, DBN and a group of resampling methods including ADASYN-DBN, SMOTE-DBN, SMOTE-borderline1-DBN, SMOTE-borderline2-DBN, SMOTE-SVM-DBN, on 58 KEEL benchmark datasets. ECS-DBN outperforms DBN and a group of resampling methods on 51 out of 58 benchmark datasets.}
	\label{tab:imbalance_benchmark_result_gmean}
	\begin{scriptsize}
		\begin{tabular}{l|ccccccc}
			\hline
			\hline
			\multirow{4}{*}{Datasets}      &  \multirow{4}{*}{ECS-DBN} & \multirow{4}{*}{DBN} & \multicolumn{5}{c}{Resampling Methods}\\\cline{4-8}\\
			& &                    & ADASYN-DBN      & SMOTE-DBN     & \begin{tabular}[c]{@{}c@{}}SMOTE -borderline1\\-DBN \end{tabular} & \begin{tabular}[c]{@{}c@{}}SMOTE-borderline2\\-DBN \end{tabular} & \begin{tabular}[c]{@{}c@{}}SMOTE-SVM\\-DBN \end{tabular} \\ \hline
			abalone9-18                                                                 & \textbf{0.7236$\pm$0.0064}        & 0.0066$\pm$0.0066 & 0.1422$\pm$0.1365          & 0.6387$\pm$0.0508          & 0.6638$\pm$0.2352 & 0.6266$\pm$0.2518 & 0.5989$\pm$0.0441          \\
			cleveland-0\_vs\_4                                                          & \textbf{0.8596$\pm$0.0180}        & 0.0222$\pm$0.0222 & 0.2681$\pm$0.2321          & 0.6411$\pm$0.2960          & 0.7689$\pm$0.1476 & 0.6946$\pm$0.3008 & 0.6620$\pm$0.3285          \\
			ecoli-0-1-3-7\_vs\_2-6                                                      & \textbf{0.9927$\pm$0.0072}        & 0.0149$\pm$0.0149 & 0.6970$\pm$0.2810          & 0.7474$\pm$0.1940          & 0.7419$\pm$0.2213 & 0.5795$\pm$0.3997 & 0.6626$\pm$0.3273          \\
			ecoli-0-1-4-6\_vs\_5                                                        & \textbf{0.8944$\pm$0.0113}        & 0.0084$\pm$0.0084 & 0.0000$\pm$0.0000          & 0.7910$\pm$0.1817          & 0.3182$\pm$0.3014 & 0.6245$\pm$0.3428 & 0.8010$\pm$0.1923          \\
			ecoli-0-1-4-7\_vs\_2-3-5-6                                                  & \textbf{0.8604$\pm$0.0176}        & 0.0189$\pm$0.0189 & 0.0639$\pm$0.0220          & 0.6336$\pm$0.3174          & 0.8902$\pm$0.1021 & 0.6303$\pm$0.3097 & 0.5635$\pm$0.4223          \\
			ecoli-0-1-4-7\_vs\_5-6                                                      & \textbf{0.9076$\pm$0.0157}        & 0.0183$\pm$0.0183 & 0.2073$\pm$0.2045          & 0.6931$\pm$0.2285          & 0.6988$\pm$0.3010 & 0.8686$\pm$0.0800 & 0.7035$\pm$0.2792          \\
			ecoli-0-1\_vs\_2-3-5                                                        & \textbf{0.9129$\pm$0.0177$^\dag$} & 0.0169$\pm$0.0169 & 0.1240$\pm$0.1215          & 0.3940$\pm$0.3252          & 0.5898$\pm$0.3323 & 0.4235$\pm$0.3268 & 0.4553$\pm$0.4246          \\
			ecoli-0-1\_vs\_5                                                            & \textbf{0.8516$\pm$0.0153}        & 0.0145$\pm$0.0145 & 0.0000$\pm$0.0000          & 0.8232$\pm$0.1246          & 0.7070$\pm$0.2753 & 0.6193$\pm$0.2982 & 0.8439$\pm$0.1487          \\
			ecoli-0-2-3-4\_vs\_5                                                        & \textbf{0.8374$\pm$0.0126}        & 0.0159$\pm$0.0159 & 0.0000$\pm$0.0000          & 0.5807$\pm$0.4091          & 0.2328$\pm$0.1821 & 0.3059$\pm$0.3039 & 0.6557$\pm$0.2635          \\
			ecoli-0-2-6-7\_vs\_3-5                                                      & \textbf{0.9192$\pm$0.0170$^\dag$} & 0.0186$\pm$0.0186 & 0.0000$\pm$0.0000          & 0.3060$\pm$0.3011          & 0.0534$\pm$0.0527 & 0.0333$\pm$0.0254 & 0.3763$\pm$0.3055          \\
			ecoli-0-3-4-6\_vs\_5                                                        & \textbf{0.9324$\pm$0.0146$^\dag$} & 0.0156$\pm$0.0156 & 0.0000$\pm$0.0000          & 0.5209$\pm$0.4542          & 0.5688$\pm$0.3637 & 0.2062$\pm$0.2060 & 0.5159$\pm$0.4543          \\
			ecoli-0-3-4-7\_vs\_5-6                                                      & \textbf{0.9065$\pm$0.0117$^\dag$} & 0.0138$\pm$0.0138 & 0.0801$\pm$0.0032          & 0.4437$\pm$0.3603          & 0.5069$\pm$0.4458 & 0.4702$\pm$0.3572 & 0.5618$\pm$0.4039          \\
			ecoli-0-3-4\_vs\_5                                                          & \textbf{0.9133$\pm$0.0134}        & 0.9098$\pm$0.0578 & 0.0000$\pm$0.0000          & 0.6126$\pm$0.3645          & 0.6592$\pm$0.2724 & 0.2383$\pm$0.1595 & 0.7176$\pm$0.2333          \\
			ecoli-0-4-6\_vs\_5                                                          & \textbf{0.9746$\pm$0.0157$^\dag$} & 0.0156$\pm$0.0156 & 0.0000$\pm$0.0000          & 0.4899$\pm$0.3660          & 0.4509$\pm$0.3661 & 0.2919$\pm$0.2219 & 0.6553$\pm$0.3237          \\
			ecoli-0-6-7\_vs\_3-5                                                        & \textbf{0.9243$\pm$0.0165$^\dag$} & 0.0169$\pm$0.0169 & 0.2106$\pm$0.2024          & 0.4873$\pm$0.3854          & 0.2074$\pm$0.1146 & 0.1794$\pm$0.1752 & 0.6440$\pm$0.3504          \\
			ecoli-0-6-7\_vs\_5                                                          & \textbf{0.8926$\pm$0.0147$^\dag$} & 0.0139$\pm$0.0139 & 0.0000$\pm$0.0000          & 0.3887$\pm$0.3800          & 0.2565$\pm$0.2451 & 0.2106$\pm$0.2069 & 0.3378$\pm$0.2949          \\
			ecoli3                                                                      & 0.7112$\pm$0.0142                 & 0.0099$\pm$0.0099 & 0.0000$\pm$0.0000          & 0.5506$\pm$0.4043          & 0.5991$\pm$0.3587 & 0.5730$\pm$0.3270 & \textbf{0.7712$\pm$0.2038} \\
			glass-0-1-2-3\_vs\_4-5-6                                                    & \textbf{0.9877$\pm$0.0084$^\dag$} & 0.0879$\pm$0.0879 & 0.1851$\pm$0.1240          & 0.5906$\pm$0.4080          & 0.1368$\pm$0.0992 & 0.0928$\pm$0.0857 & 0.6528$\pm$0.3060          \\
			glass-0-1-4-6\_vs\_2                                                        & \textbf{0.9266$\pm$0.0041$^\dag$} & 0.0000$\pm$0.0000 & 0.0000$\pm$0.0000          & 0.0919$\pm$0.0904          & 0.1256$\pm$0.1057 & 0.0416$\pm$0.0316 & 0.0000$\pm$0.0000          \\
			glass-0-1-5\_vs\_2                                                          & \textbf{0.9837$\pm$0.0044$^\dag$} & 0.0000$\pm$0.0000 & 0.0000$\pm$0.0000          & 0.1440$\pm$0.1413          & 0.1048$\pm$0.1025 & 0.0224$\pm$0.0107 & 0.0000$\pm$0.0000          \\
			glass-0-1-6\_vs\_2                                                          & \textbf{0.7030$\pm$0.0186$^\dag$} & 0.0000$\pm$0.0000 & 0.0000$\pm$0.0000          & 0.0814$\pm$0.0734          & 0.2035$\pm$0.1158 & 0.1666$\pm$0.1261 & 0.0000$\pm$0.0000          \\
			glass-0-1-6\_vs\_5                                                          & \textbf{0.6785$\pm$0.0058}        & 0.0066$\pm$0.0066 & 0.1201$\pm$0.0535          & 0.4598$\pm$0.4051          & 0.3747$\pm$0.2317 & 0.5874$\pm$0.3625 & 0.0000$\pm$0.0000          \\
			glass-0-4\_vs\_5                                                            & \textbf{0.9837$\pm$0.0083}        & 0.8943$\pm$0.0860 & 0.1336$\pm$0.0977          & 0.1665$\pm$0.1558          & 0.0930$\pm$0.0841 & 0.0310$\pm$0.0280 & 0.2641$\pm$0.2267          \\
			glass-0-6\_vs\_5                                                            & \textbf{0.9486$\pm$0.0174$^\dag$} & 0.0187$\pm$0.0187 & 0.0599$\pm$0.0396          & 0.0000$\pm$0.0000          & 0.2694$\pm$0.2583 & 0.2103$\pm$0.1474 & 0.4218$\pm$0.3341          \\
			glass0                                                                      & \textbf{0.6785$\pm$0.0362$^\dag$} & 0.0327$\pm$0.0327 & 0.0000$\pm$0.0000          & 0.3342$\pm$0.2762          & 0.1932$\pm$0.0576 & 0.0160$\pm$0.0106 & 0.2918$\pm$0.2125          \\
			glass1                                                                      & \textbf{0.7641$\pm$0.0139$^\dag$} & 0.0151$\pm$0.0151 & 0.0000$\pm$0.0000          & 0.1163$\pm$0.1053          & 0.1322$\pm$0.1170 & 0.1103$\pm$0.0289 & 0.1810$\pm$0.1367          \\
			glass2                                                                      & \textbf{0.5105$\pm$0.0636$^\dag$} & 0.0000$\pm$0.0000 & 0.0000$\pm$0.0000          & 0.0967$\pm$0.0908          & 0.1625$\pm$0.0684 & 0.0804$\pm$0.0705 & 0.0000$\pm$0.0000          \\
			glass4                                                                      & \textbf{0.9486$\pm$0.0261}        & 0.0237$\pm$0.0237 & 0.0594$\pm$0.0252          & 0.5909$\pm$0.4081          & 0.5393$\pm$0.4115 & 0.5503$\pm$0.3909 & 0.1234$\pm$0.0606          \\
			glass5                                                                      & \textbf{0.9951$\pm$0.0019$^\dag$} & 0.0017$\pm$0.0017 & 0.0000$\pm$0.0000          & 0.3332$\pm$0.3302          & 0.4054$\pm$0.3934 & 0.4876$\pm$0.3948 & 0.0000$\pm$0.0000          \\
			glass6                                                                      & \textbf{0.9354$\pm$0.0644}        & 0.0624$\pm$0.0624 & 0.1188$\pm$0.0641          & 0.6041$\pm$0.3170          & 0.5225$\pm$0.4514 & 0.6703$\pm$0.2490 & 0.6130$\pm$0.3232          \\
			haberman                                                                    & \textbf{0.5105$\pm$0.0224}        & 0.0248$\pm$0.0248 & 0.0000$\pm$0.0000          & 0.1400$\pm$0.1295          & 0.0746$\pm$0.0449 & 0.2895$\pm$0.2601 & 0.0959$\pm$0.0022          \\
			iris0                                                                       & \textbf{0.9450$\pm$0.0118}        & 0.2006$\pm$0.2006 & 0.0000$\pm$0.0000          & 0.8882$\pm$0.1043          & 0.1000$\pm$0.0162 & 0.1957$\pm$0.1128 & 0.1975$\pm$0.1163          \\
			\begin{tabular}[c]{@{}l@{}}led7digit-0-2-4-5-6-7\\ -8-9\_vs\_1\end{tabular} & \textbf{0.9951$\pm$0.0043}        & 0.9731$\pm$0.0238 & 0.7109$\pm$0.0184          & 0.8248$\pm$0.0062          & 0.8713$\pm$0.1263 & 0.8218$\pm$0.1001 & 0.8980$\pm$0.0048          \\
			new-thyroid1                                                                & \textbf{0.7379$\pm$0.0787$^\dag$} & 0.0784$\pm$0.0784 & 0.0000$\pm$0.0000          & 0.3671$\pm$0.3113          & 0.1188$\pm$0.1187 & 0.0202$\pm$0.0139 & 0.1721$\pm$0.1591          \\
			newthyroid2                                                                 & \textbf{0.8992$\pm$0.0841$^\dag$} & 0.0812$\pm$0.0812 & 0.0000$\pm$0.0000          & 0.2428$\pm$0.1940          & 0.1434$\pm$0.0648 & 0.0000$\pm$0.0000 & 0.1773$\pm$0.1409          \\
			page-blocks-1-3\_vs\_4                                                      & \textbf{0.9837$\pm$0.0159}        & 0.0242$\pm$0.0242 & 0.6504$\pm$0.2320          & 0.5591$\pm$0.3259          & 0.8324$\pm$0.0776 & 0.7323$\pm$0.2629 & 0.7211$\pm$0.0211          \\
			page-blocks0                                                                & \textbf{0.8649$\pm$0.0944}        & 0.8595$\pm$0.0123 & 0.8013$\pm$0.0174          & 0.8593$\pm$0.0150          & 0.8044$\pm$0.0156 & 0.6339$\pm$0.2395 & 0.8495$\pm$0.0294          \\
			pima                                                                        & \textbf{0.9735$\pm$0.0204$^\dag$} & 0.0413$\pm$0.0413 & 0.2468$\pm$0.1941          & 0.5283$\pm$0.2088          & 0.3099$\pm$0.2907 & 0.2580$\pm$0.2328 & 0.2581$\pm$0.2282          \\
			segment0                                                                    & \textbf{0.8678$\pm$0.0494}        & 0.0503$\pm$0.0503 & 0.9261$\pm$0.0164          & 0.6752$\pm$0.3240          & 0.8240$\pm$0.1758 & 0.7582$\pm$0.2410 & 0.9358$\pm$0.0141          \\
			shuttle-c0-vs-c4                                                            & \textbf{0.9864$\pm$0.0132}        & 0.9837$\pm$0.0000 & 0.0000$\pm$0.0000          & 0.8000$\pm$0.1216          & 0.0189$\pm$0.0098 & 0.0378$\pm$0.0297 & 0.6984$\pm$0.2819          \\
			shuttle-c2-vs-c4                                                            & \textbf{0.7584$\pm$0.0045$^\dag$} & 0.0040$\pm$0.0040 & 0.0775$\pm$0.0449          & 0.4249$\pm$0.3756          & 0.2677$\pm$0.2328 & 0.4859$\pm$0.2866 & 0.0000$\pm$0.0000          \\
			vehicle0                                                                    & \textbf{0.9735$\pm$0.0196$^\dag$} & 0.0669$\pm$0.0669 & 0.6686$\pm$0.2367          & 0.6940$\pm$0.2683          & 0.6598$\pm$0.3107 & 0.5708$\pm$0.2897 & 0.6381$\pm$0.3452          \\
			vowel0                                                                      & 0.5136$\pm$0.0231                 & 0.0224$\pm$0.0224 & 0.8880$\pm$0.0780          & \textbf{0.9052$\pm$0.0207} & 0.7299$\pm$0.2572 & 0.8465$\pm$0.0189 & 0.8968$\pm$0.0448          \\
			wisconsin                                                                   & \textbf{0.9864$\pm$0.0114}        & 0.1327$\pm$0.1327 & 0.9694$\pm$0.0035          & 0.8716$\pm$0.1063          & 0.8761$\pm$0.1084 & 0.8702$\pm$0.1058 & 0.9748$\pm$0.0150          \\
			yeast-0-2-5-6\_vs\_3-7-8-9                                                  & 0.7584$\pm$0.0301                 & 0.0341$\pm$0.0341 & 0.3035$\pm$0.2881          & 0.6602$\pm$0.3277          & 0.6855$\pm$0.2461 & 0.7766$\pm$0.0069 & \textbf{0.8042$\pm$0.1383} \\
			yeast-0-2-5-7-9\_vs\_3-6-8                                                  & \textbf{0.9545$\pm$0.0436}        & 0.0489$\pm$0.0489 & 0.5075$\pm$0.2968          & 0.6991$\pm$0.2477          & 0.5131$\pm$0.3533 & 0.5067$\pm$0.3512 & 0.8030$\pm$0.1911          \\
			yeast-0-3-5-9\_vs\_7-8                                                      & \textbf{0.8822$\pm$0.0164$^\dag$} & 0.0180$\pm$0.0180 & 0.0000$\pm$0.0000          & 0.4785$\pm$0.1819          & 0.3552$\pm$0.2588 & 0.2772$\pm$0.2208 & 0.5127$\pm$0.1803          \\
			yeast-0-5-6-7-9\_vs\_4                                                      & \textbf{0.8944$\pm$0.0144}        & 0.0161$\pm$0.0161 & 0.0552$\pm$0.0254          & 0.8093$\pm$0.0301          & 0.8197$\pm$0.0608 & 0.6566$\pm$0.2634 & 0.7896$\pm$0.1792          \\
			yeast-1-2-8-9\_vs\_7                                                        & \textbf{0.6975$\pm$0.0003}        & 0.0004$\pm$0.0004 & 0.0000$\pm$0.0000          & 0.5120$\pm$0.3539          & 0.6440$\pm$0.2282 & 0.4730$\pm$0.2764 & 0.0000$\pm$0.0000          \\
			yeast-1-4-5-8\_vs\_7                                                        & \textbf{0.8101$\pm$0.0031$^\dag$} & 0.0000$\pm$0.0000 & 0.0000$\pm$0.0000          & 0.2794$\pm$0.2122          & 0.5847$\pm$0.2132 & 0.3642$\pm$0.2674 & 0.0000$\pm$0.0000          \\
			yeast-1\_vs\_7                                                              & 0.6323$\pm$0.0081                 & 0.0086$\pm$0.0086 & 0.0000$\pm$0.0000          & 0.6678$\pm$0.0673          & 0.6861$\pm$0.0943 & 0.5891$\pm$0.2111 & \textbf{0.7442$\pm$0.2518} \\
			yeast-2\_vs\_4                                                              & \textbf{0.8822$\pm$0.0279}        & 0.8818$\pm$0.0180 & 0.0225$\pm$0.0213          & 0.8668$\pm$0.0308          & 0.4358$\pm$0.3818 & 0.4821$\pm$0.2760 & 0.7102$\pm$0.2744          \\
			yeast-2\_vs\_8                                                              & \textbf{0.8944$\pm$0.0301$^\dag$} & 0.0303$\pm$0.0303 & 0.0519$\pm$0.0341          & 0.5792$\pm$0.3053          & 0.2663$\pm$0.2344 & 0.2160$\pm$0.2065 & 0.5753$\pm$0.2132          \\
			yeast1                                                                      & \textbf{0.6975$\pm$0.0074$^\dag$} & 0.0331$\pm$0.0331 & 0.0454$\pm$0.0049          & 0.2532$\pm$0.2297          & 0.2979$\pm$0.2633 & 0.2359$\pm$0.2170 & 0.5126$\pm$0.2635          \\
			yeast3                                                                      & \textbf{0.8101$\pm$0.0174}        & 0.7992$\pm$0.0238 & 0.0000$\pm$0.0000          & 0.6761$\pm$0.2563          & 0.5040$\pm$0.4158 & 0.4681$\pm$0.3152 & 0.3880$\pm$0.3335          \\
			yeast4                                                                      & 0.6323$\pm$0.0369                 & 0.5062$\pm$0.1069 & 0.5666$\pm$0.4055          & 0.7744$\pm$0.1721          & 0.8295$\pm$0.1703 & 0.8982$\pm$0.0024 & \textbf{0.9289$\pm$0.0032} \\
			yeast5                                                                      & 0.8537$\pm$0.0201                 & 0.0099$\pm$0.0099 & \textbf{0.9654$\pm$0.0007} & 0.7719$\pm$0.2068          & 0.7815$\pm$0.1819 & 0.9269$\pm$0.0014 & 0.8734$\pm$0.1169          \\
			yeast6                                                                      & 0.7188$\pm$0.0360                 & 0.0083$\pm$0.0083 & 0.5551$\pm$0.3778          & 0.7979$\pm$0.1804          & 0.8425$\pm$0.1560 & 0.7316$\pm$0.2478 & \textbf{0.9530$\pm$0.0020}
			\\
			\hline    
			Average Rank                                                     & \textbf{1.4310} & 5.8103 & 5.8534                                              & 3.4483                                               & 3.6207                                                              & 4.3879                                                              & 3.4483                                                      \\ 
			Win-lose-draw& - & 51-0-7 & 52-0-6 & 43-3-12 &	45-2-11    & 44-4-10 & 44-2-12    \\
			Holm-Post hoc Test & - & \textbf{1.75303E-11}	 & \textbf{4.45971E-11}	  & \textbf{3.65941E-10}	   & \textbf{2.72630E-10}	    & \textbf{2.59525E-10}	     & \textbf{4.10284E-09}	     \\
			\hline                                              
		\end{tabular}
		\\$\dag$ indicates that the difference between the proposed algorithm and all other compared algorithm is statistically significant using pair-wised Wilcoxon rank sum test at the $5\%$ significance level.
	\end{scriptsize}
\end{table*}

\end{document}